\documentclass[10pt,twocolumn,letterpaper]{article}

\usepackage[pagenumbers]{wacv}

\definecolor{wacvblue}{rgb}{0.21,0.49,0.74}
\usepackage[pagebackref,breaklinks,colorlinks,allcolors=wacvblue]{hyperref}

\def\wacvPaperID{368} 
\def\confName{WACV}
\def\confYear{2026}

\title{OW-Rep: Open World Object Detection with Instance Representation Learning}

\author{
    Sunoh Lee\textsuperscript{1}\thanks{Equal contribution. Work primarily conducted while at Agency for Defense Development. \newline
    \makebox[1.0em][l]{}Project website: \href{https://sunohlee.github.io/OW-Rep/}{\tt\footnotesize https://sunohlee.github.io/OW-Rep/}} \quad
    Minsik Jeon\textsuperscript{2}\footnotemark[1] \quad
    Jihong Min\textsuperscript{3} \quad
    Junwon Seo\textsuperscript{2} \\[0.3em]
    \textsuperscript{1}KAIST \quad
    \textsuperscript{2}Carnegie Mellon University \quad
    \textsuperscript{3}Agency for Defense Development\\
    {\tt\small sunoh0131@kaist.ac.kr} \quad
    {\tt\small \{minsikj, junwonse\}@andrew.cmu.edu} \quad
    {\tt\small happymin77@gmail.com}
}

\usepackage{pifont}
\usepackage{multirow}
\usepackage{colortbl}
\usepackage{graphicx}
\usepackage{xcolor}
\usepackage{booktabs}
\usepackage{array}
\usepackage[accsupp]{axessibility}  
\usepackage{float}
\usepackage{algorithm}
\usepackage{algpseudocode}
\usepackage{etoc}

\begin{document}
\etocdepthtag.toc{mtoc}
\maketitle

\begin{abstract}
\hspace{0.15in} Open World Object Detection~(OWOD) addresses realistic scenarios where unseen object classes emerge, enabling detectors trained on known classes to detect unknown objects and incrementally incorporate the knowledge they provide. While existing OWOD methods primarily focus on detecting unknown objects, they often overlook the rich semantic relationships between detected objects, which are essential for scene understanding and applications in open-world environments (e.g., open-world tracking and novel class discovery). In this paper, we extend the OWOD framework to jointly detect unknown objects and learn semantically rich instance embeddings, enabling the detector to capture fine-grained semantic relationships between instances. To this end, we propose two modules that leverage the rich and generalizable knowledge of Vision Foundation Models~(VFMs) and can be integrated into open-world object detectors. First, the Unknown Box Refine Module uses instance masks from the Segment Anything Model to accurately localize unknown objects. The Embedding Transfer Module then distills instance-wise semantic similarities from VFM features to the detector's embeddings via a relaxed contrastive loss, enabling the detector to learn a semantically meaningful and generalizable instance feature. Extensive experiments show that our method significantly improves both unknown object detection and instance embedding quality, while also enhancing performance in downstream tasks such as open-world tracking.

\end{abstract}

\vspace{-0.20in}
\section{Introduction}
\vspace{-0.10in}
\label{sec:Introduction}
\hspace{0.15in} 
Humans naturally detect and localize unfamiliar objects, drawing on associations with similar ones to comprehend them. While deep learning-based methods have made significant improvements in perception, object detectors still struggle to detect objects beyond the training distribution. To operate reliably in real-world environments, detectors should be able to detect and identify both the previously known and novel unknown objects. Open World Object Detection~(OWOD)~\cite{joseph2021towards}, recently introduced, enhanced the detectors' practicality and reliability by enabling the detection of unknown objects not labeled in the training set. 
The framework then involves a human annotator to label detected unknowns and incrementally trains the detector on this new data, enabling continuous adaptation in open-world environments.

\begin{figure}[t]
\begin{center}
\includegraphics[width=1.0\linewidth]{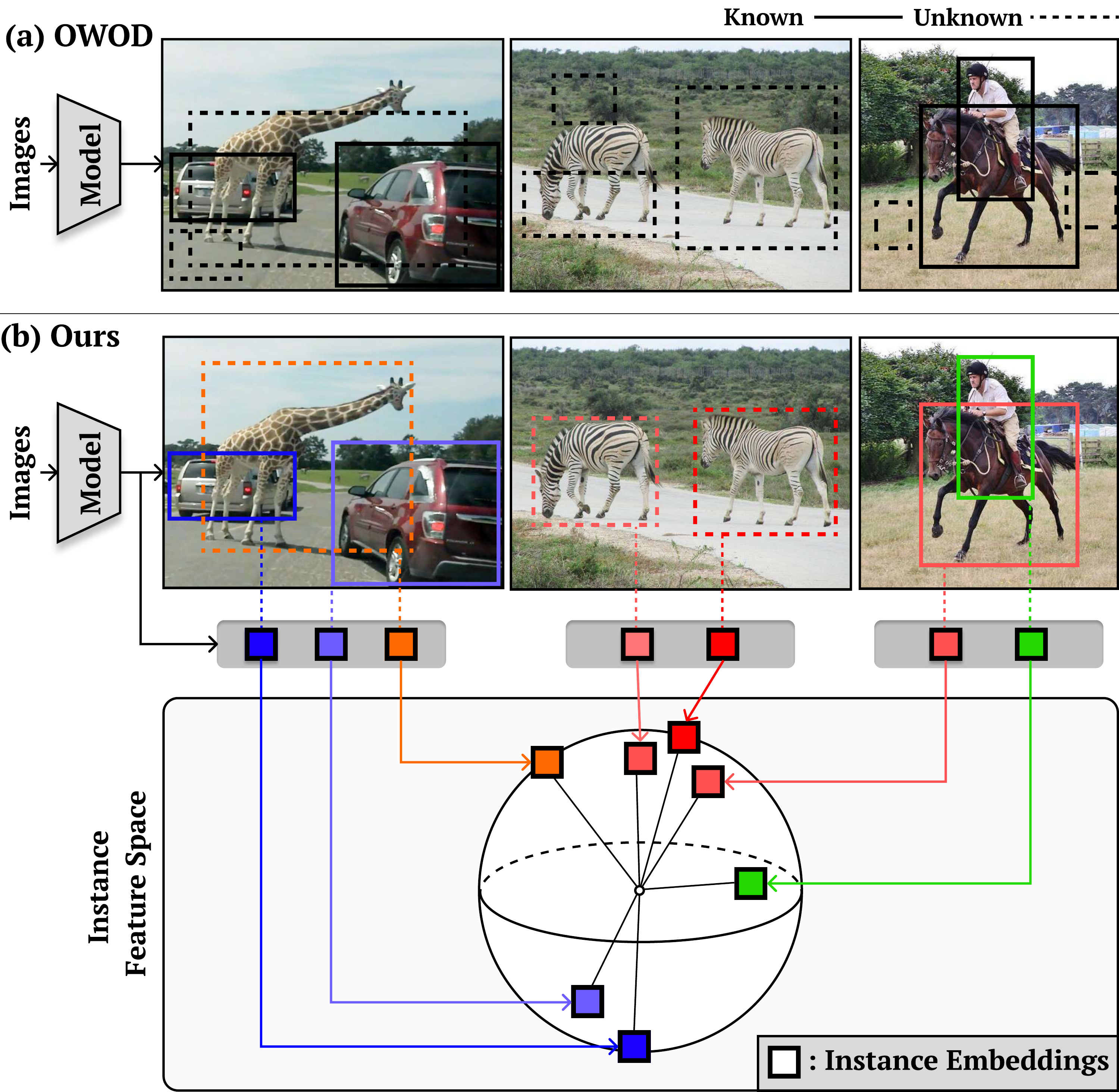}
\end{center}
\caption{
We propose a method for training an open-world object detector that not only detects unknown objects but also learns semantically rich feature embeddings that capture meaningful inter-object relationships. Existing OWOD methods~\cite{joseph2021towards, gupta2022ow, zohar2023prob, doan_2024_HypOW} primarily focus on detecting unknown objects but overlook the semantic relationships between proposals. Our approach explicitly captures these relationships by enhanced instance embeddings through VFM~\cite{oquab2023dinov2, kirillov2023segment} distillation, while also improving unknown object detection.
}
\label{fig:concept}
\end{figure}

To achieve a comprehensive understanding of a scene in open-world environments, it is crucial to not only detect both known and unknown objects but also to learn the finer relationships between them. These relationships play a key role in diverse applications of detectors where predefined information about novel classes is unavailable. For example, tasks such as object tracking~\cite{liu2022opening, pang2021quasi, wojke2017simple}, path planning~\cite{honda2024replan}, and class discovery~\cite{wu2022uc, fomenko2022learning} in open-world scenarios heavily rely on semantic similarity between detected objects for association and obstacle identification. However, existing OWOD approaches primarily focus on learning generalized object representations across diverse categories, often overlooking these finer relationships and failing to capture meaningful semantic information. As a result, methods relying on pretrained detector features to infer relationships between known and unknown objects often struggle to work effectively~\cite{rambhatla2021pursuit, hayes2024pandas}.

Since existing OWOD methods lack supervision to learn rich features of unknown objects, a few studies have tried to enhance the feature quality by self-supervised learning using unknown proposals~\cite{wu2022uc, fomenko2022learning,zheng2022towards}. However, the inaccurate proposals from detectors hinder the methods of learning robust features. Moreover, these methods mainly focus on learning representations of unknown objects present in the training data, which we refer to as \textit{known-unknown}s. In real-world deployment, systems may also encounter \textit{unknown-unknown} objects that were never observed during training~\cite{dhamija2020overlooked}. Therefore, the learned feature space should be able to embed both of them properly based on their semantics for robust operation in open-world scenarios.

In this paper, we extend the OWOD framework to simultaneously detect unknown objects and extract semantically rich features (Fig.~\ref{fig:concept}) by introducing modules that can be jointly trained with a baseline open-world object detector.
To enable accurate localization and the learning of semantically rich feature embeddings, our approach leverages the knowledge of Vision Foundation Models (VFMs). Specifically, segmentation masks from Segment Anything Model~(SAM)~\cite{kirillov2023segment} are employed to supervise the localization of unknown objects, while the detector's instance features are enriched by distilling the similarity relations derived from VFM’s pixel-level features~\cite{oquab2023dinov2}. This distillation is performed using a relaxed contrastive loss~\cite{kim2021embedding}, which provides a rich supervisory signal that enables the learning of a generalizable feature space. Our contribution is the joint integration of box refinement and relaxed contrastive transfer, which allows the detector to learn semantically rich features despite the lack of direct supervision and the ambiguity of class boundaries in open-world scenarios.

We demonstrate the effectiveness of our method through extensive experiments, including downstream applications. Compared to prior self-supervised approaches, our method significantly improves unknown object detection and instance embedding quality. Performance gains across various baseline open-world detectors further validate its strong compatibility. Evaluation with a novel benchmark with \textit{unknown-unknown}s shows the generalizability of our method in properly embedding both the \textit{known-unknown} and \textit{unknown-unknown}s. Finally, through an application to open-world tracking, we demonstrate that learning semantically rich instance features benefits downstream tasks. 


\section{Related Works}
\vspace{-0.07in}
\label{sec:Related_Works}

\noindent\textbf{Open World Object Detection.}
The Open World Object Detection~(OWOD) framework is introduced to detect objects in realistic environments where previously unseen objects may appear, allowing the detector to progressively evolve by incorporating newly detected objects~\cite{joseph2021towards}. 
To identify novel objects not labeled in the training set, existing methods learn a generalized understanding of \textit{object} by utilizing the knowledge from labeled, known objects~\cite{joseph2021towards, gupta2022ow, zohar2023prob, zohar2023open, kim2022learning, wang2023detecting, wang2023random, ma2021annealing, liang2023unknown, doan_2024_HypOW, sun2024exploring}. 
Some methods directly classify the proposals as either known or unknown using energy functions or activation scores~\cite{joseph2021towards,gupta2022ow}, while \textit{PROB}~\cite{zohar2023prob} learns an objectness score based on the distribution of known object features. \textit{Hyp-OW}~\cite{doan_2024_HypOW} further grouped known objects into superclasses and computed objectness based on these groupings. However, as these methods focus on learning a generalized concept of objects, they struggle to distinguish between different unknown objects.

For OWOD to be applicable in diverse real-world scenarios, it must not only detect unknown objects but also capture detailed relationships between them. In particular, open-world tracking~\cite{liu2022opening, pang2021quasi} associates objects across frames based on semantic similarities between pairwise proposals. Similarly, path planning and mapping often require determining which known class an unknown object mostly resembles to predict its motion or refine maps~\cite{sodano2024open}.
Novel class discovery also depends on proposal similarities to identify new categories~\cite {wu2022uc, fomenko2022learning, hayes2024pandas, rambhatla2021pursuit}.  Although some methods acquire such relationships by clustering fixed detector features~\cite{hayes2024pandas,rambhatla2021pursuit}, these groupings tend to be coarse, and the features themselves lack meaningful information. To effectively capture detailed relationships, the detector must learn semantically rich instance features directly.

\begin{figure*}[t]
\begin{center}
\includegraphics[width=1.0\linewidth]{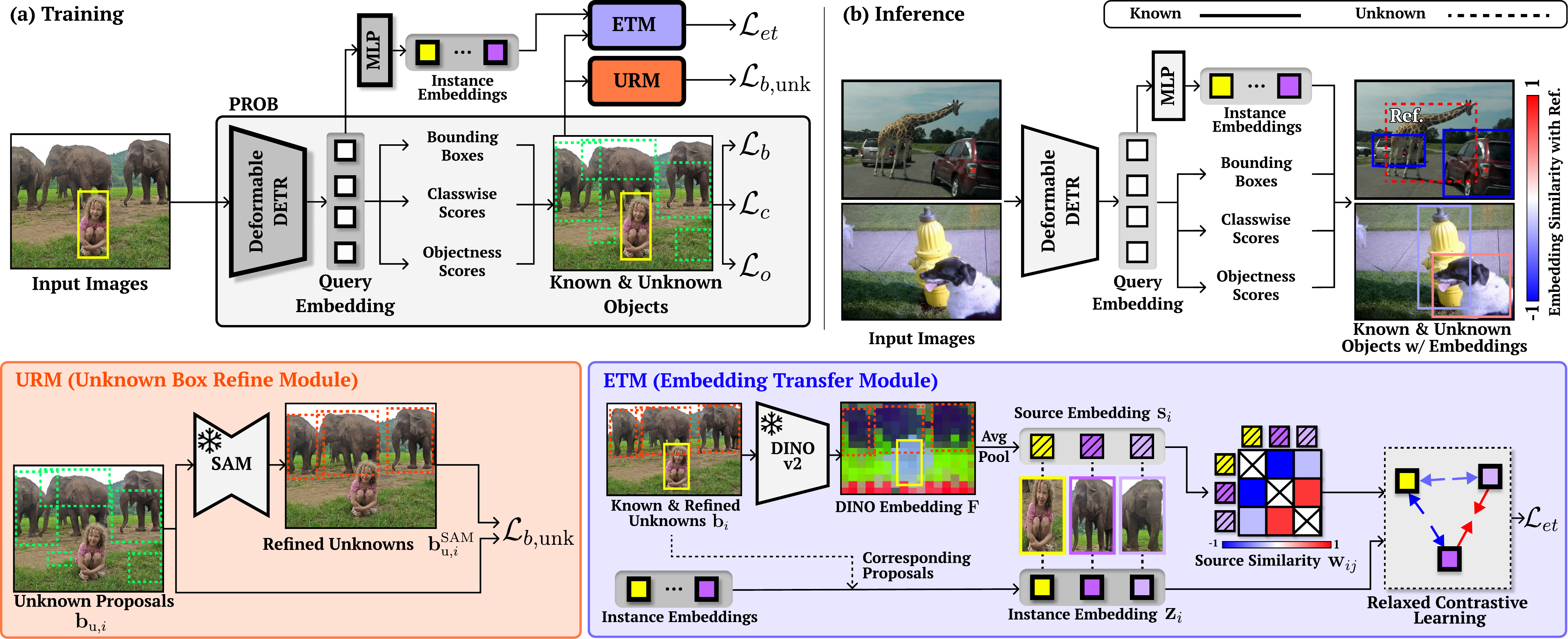}
\end{center}
\vspace{-0.20in}
\caption{
\textbf{Overall Architecture of the Proposed Method.} Our method extends OWOD by not only detecting unknown objects but also extracting semantically rich features. We adopt \textit{PROB}~\cite{zohar2023prob} as the baseline detector, but any other open-world object detector can be utilized as a baseline.
During training, the known and unknown proposals from PROB, with corresponding instance embeddings, are fed into the proposed modules. The \textit{Unknown Box Refine Module} improves the localization of unknown objects by treating refined unknown boxes from SAM~\cite{kirillov2023segment} as pseudo ground truth. The \textit{Embedding Transfer Module} extracts source embeddings by average pooling DINOv2~\cite{oquab2023dinov2} features within the refined unknown and known proposals. Pairwise similarities between source embeddings are then computed and used as weights for the relaxed contrastive loss~\cite{kim2021embedding}, controlling the attraction and repulsion between instance embeddings. At inference, the detector generates semantically rich instance embeddings, capturing fine-grained relationships between detected proposals.
}
\label{fig:model}
\vspace{-0.20in}
\end{figure*}

\noindent\textbf{Self-Supervised Representation Learning.}
Due to the lack of supervision for unknown objects, open-world detection methods have employed self-supervised learning techniques~\cite{chen2020simple, oord2018representation,he2020momentum,kim2021embedding, passalis2018learning} to obtain rich feature embeddings for each instance. \textit{OSODD}~\cite{zheng2022towards} trains a separate proposal embedding network through contrastive learning on augmented pairs of cropped proposals~\cite{he2020momentum}. In contrast, some approaches directly learn instance-level features from the detector itself. \textit{UC-OWOD}~\cite{wu2022uc} strengthens relationships between highly similar or dissimilar proposals through contrastive learning, while \textit{RNCDL}~\cite{fomenko2022learning} clusters all proposals into a predefined number of classes and treats those within the same cluster as positive pairs~\cite{asano2019self, cuturi2013sinkhorn}. However, the effectiveness of these methods depends heavily on the quality of unknown proposals, which is often degraded by inaccurate boxes and high false positive rates in open-world detectors~\cite{joseph2021towards,gupta2022ow,zohar2023prob}. Furthermore, these methods use binary labels for contrastive loss to enforce unknown class separation, thus failing to capture detailed relationships between instances~\cite{kim2021embedding}. To learn richer and more robust features for OWOD, a more informative and accurate supervisory signal is required.

\noindent\textbf{Foundation Model for Generalizing Object Detection.}
Recent efforts have explored using foundation models for unknown object detection due to their strong generalization capabilities in previously unseen scenarios~\cite{minderer2022simple, zohar2023open, cheng2024yolo, cao2024coda, wang2024ov, kim2024region,li2022grounded,liu2024grounding}. However, these models are computationally expensive, as they rely on large-scale image-text or instance-text datasets and often require models during inference. To mitigate this, several approaches focus on distilling knowledge from foundation models into object detectors~\cite{faber2024leveraging, li2024matching,he2023usd, gu2021open, kuo2022f, wu2023aligning, li2023distilling}. For example, instance masks generated by SAM can improve unknown object localization~\cite{he2023usd}. Some methods also distill feature from vision-language models to learn unknown object attributes and improve detection performance~\cite{gu2021open, kuo2022f, wu2023aligning, li2023distilling,wang2023open,wang2023learning,thawakar2025video,zang2022open}. However, these approaches can only detect unknown objects that match predefined text prompts. As a result, they fail to detect truly novel objects outside the prompt set and cannot support incremental learning based on the detection results of such objects.


\vspace{-0.05in}
\section{Methods}
\label{sec:Method}

\subsection{Problem Formulation}
\vspace{-0.05in}
\hspace{0.15in} We extend the open-world object detection setting by not only detecting unknown objects and incrementally incorporating them into the training data, but also by simultaneously extracting semantically rich features for all detected objects. At time step $t$, the dataset $D^t=\{I^t,y^t\}$ consists of input images $I^t$ and the corresponding labels $y^t$ , where only objects from $K^t$ known classes are annotated. The detector's objective is to detect both objects from the $K^t$ known classes and objects from unknown classes. Simultaneously, our method extracts semantically rich features for detected objects, embedding semantically similar objects close to each other in feature space.

\subsection{Overall Architecture}
\hspace{0.15in} 
The overall architecture of the proposed method is illustrated in Fig~\ref{fig:model}.
We adopt \textit{PROB}~\cite{zohar2023prob} as the baseline detector and extend it through our introduced modules. Note that any other open-world object detector can also be utilized as a baseline detector. Built on deformable DETR~\cite{zhu2020deformable}, \textit{PROB} generates query embeddings that encapsulate information about each predicted object. These embeddings are then processed to compute the bounding boxes and classification scores, where each proposals are classified as either a known object or \textit{background}. Known proposals are matched with ground truth objects via Hungarian matching, and their regression and classification loss, $\mathcal{L}_{b}$ and $\mathcal{L}_{c}$~\cite{zohar2023prob}, are computed. For \textit{background} proposals, unknown objects are identified based on an objectness score, representing the probability that a proposal corresponds to an object. This score is learned through the objectness loss, $\mathcal{L}_{o}$~\cite{zohar2023prob}.

Based on \textit{PROB}, we propose two modules that are jointly trained with the detector in an end-to-end manner to distill rich VFM features, enabling accurate object detection and semantically rich feature extraction. 
First, the \textit{Unknown Box Refine Module} improves the localization of unknown objects by leveraging instance masks generated by SAM~\cite{kirillov2023segment}. Then, the \textit{Embedding Transfer Module} distills the instance-wise semantic similarities obtained from the VFM pixel-level features~\cite{oquab2023dinov2} to enhance the detector's instance embeddings. Both modules are used only during training and discarded at inference.

\subsection{Unknown Box Refine Module}
\hspace{0.15in} 
\label{sec:URM}
The unavailability of bounding box annotations for unknown objects poses a significant challenge for accurate localization in open-world object detection. Inaccurate boxes and numerous false positives further hinder methods that learn instance-wise embeddings, as these methods heavily rely on predicted proposals. To enhance both detection accuracy and feature quality, improving the localization of unknown objects is essential.

We employ SAM's high-quality instance masks~\cite{kirillov2023segment} to supervise the localization of unknown objects and obtain accurate bounding boxes. Given an input image $I$, we select the top-$k$ proposals with the highest objectness scores, excluding those classified as known objects. The bounding boxes of these proposals, $\mathbf{B}_{u}=\{\mathbf{b}_{\text{u},i}\}_{i=1}^{k}$, serve as prompts for SAM to generate refined unknown boxes, $\mathbf{b}_{\text{u},i}^\text{SAM}$. To filter out false positives, only refined boxes with an Intersection over Union~(IoU) above a threshold $\kappa$ with their corresponding original boxes are retained. The regression head is then trained using these refined boxes as ground truth, and the unknown regression loss calculated as follows: 
\begin{equation}
    \mathcal{L}_{b,\text{unk}}=
    \frac{1}{\vert U\vert}\sum_{i\in U}\lVert\mathbf{b}_{\text{u},i} - \mathbf{b}_{\text{u},i}^{\text{SAM}}\rVert_{1}
    -\text{GIoU}(\mathbf{b}_{\text{u},i}, \mathbf{b}_{\text{u},i}^{\text{SAM}}),
\end{equation}
where $U$ denotes the set of indices corresponding to the retained boxes with sufficiently high overlap, and GIoU represents the Generalized Intersection over Union~\cite{rezatofighi2019generalized}.


\begin{table*}[htbp]
\renewcommand\arraystretch{1.25}
\centering
\vspace{-0.15in}
\caption{\textbf{Quantitative results on OWOD split.}  
We compared our method with the other approaches that simultaneously detect unknown objects and learn instance embeddings, reporting performance on Unknown Recall, Recall@1, and known mAP. 
Our method outperforms others in both unknown object detection~(Unknown Recall) and feature embedding quality~(Recall@1) while maintaining competitive performance on known objects~(Known mAP), demonstrating its effectiveness for OWOD and instance representation learning. Note that known mAP is calculated over all known objects. Since all classes are known in Task 4, Unknown Recall and Recall@1 are not reported.
} 
\label{tab:owod_results}
\scriptsize
\resizebox{\textwidth}{!}{
\begin{tabular}{l| ccc| ccc |ccc | c}\toprule
\multicolumn{1}{c|}{\textbf{Task IDs $\rightarrow$}}& \multicolumn{3}{c|}{\textbf{Task 1}}& \multicolumn{3}{c|}{\textbf{Task 2}} & \multicolumn{3}{c|}{\textbf{Task 3}}  & \multicolumn{1}{c}{\textbf{Task 4}} \\
\midrule
\multicolumn{1}{c|}{}   & \multicolumn{1}{c}{ \cellcolor[HTML]{fcfce0}Unknown}  & \cellcolor[HTML]{f4cccc}{Recall} & \cellcolor[HTML]{e4e5fa}Known &   \multicolumn{1}{c}{ \cellcolor[HTML]{fcfce0}Unknown} & \multicolumn{1}{c}{\cellcolor[HTML]{f4cccc}{Recall}} & \multicolumn{1}{c|}{\cellcolor[HTML]{e4e5fa}Known} &  \multicolumn{1}{c}{ \cellcolor[HTML]{fcfce0}Unknown} & \multicolumn{1}{c}{\cellcolor[HTML]{f4cccc}{Recall}} & \multicolumn{1}{c|}{\cellcolor[HTML]{e4e5fa}Known}   &\multicolumn{1}{c}{\cellcolor[HTML]{e4e5fa}Known}  \\
\multicolumn{1}{c|}{\multirow{-2}{*}{Metrics $\rightarrow$}}   & \cellcolor[HTML]{fcfce0} Recall & \cellcolor[HTML]{f4cccc} @1 &  \cellcolor[HTML]{e4e5fa}mAP & \cellcolor[HTML]{fcfce0}Recall& \cellcolor[HTML]{f4cccc} @1 & \multicolumn{1}{c|}{\cellcolor[HTML]{e4e5fa}mAP}& \cellcolor[HTML]{fcfce0}Recall  &  \cellcolor[HTML]{f4cccc} @1 &\multicolumn{1}{c|}{\cellcolor[HTML]{e4e5fa}mAP}    & \multicolumn{1}{c}{\cellcolor[HTML]{e4e5fa}mAP}\\

 \midrule

\textit{PROB}\cite{zohar2023prob} & 18.84 & 4.42 &  \textbf{58.98} &  16.04 & 3.79 & 43.76 & 17.28 & 6.69 & 35.81 & \textbf{31.72}  \\

\textit{OSODD}\cite{zheng2022towards} & 18.84 & 4.85 &  \textbf{58.98} &  16.04 & 3.99 & 43.76 & 17.28 & 6.57 & 35.81 & \textbf{31.72}  \\
\textit{UC-OWOD}\cite{wu2022uc} &  12.87 &  3.84 &  45.11 & 10.04 & 2.85 & 34.04 & 13.31 & 4.83 & 29.30 & \textbf{31.72} \\
\textit{RNCDL}\cite{fomenko2022learning} & 18.93 & 1.41 & 57.51 & 15.13 & 1.38 & 43.90 & 17.35 & 2.82 & 35.54 & 31.49 \\

 \multirow{1}{*}{\textbf{\textit{Ours}}}   &\textbf{30.56}    &\textbf{11.69} & 58.89 &  \textbf{26.74}& \textbf{10.19} & \textbf{45.00}  &\textbf{28.21} & \textbf{13.25} & \textbf{36.31} &31.46\\ 

\bottomrule 
\end{tabular}}
\vspace{-0.1in}
\end{table*}


\begin{figure*}[th]
\begin{center}
\includegraphics[width=1.0\linewidth]{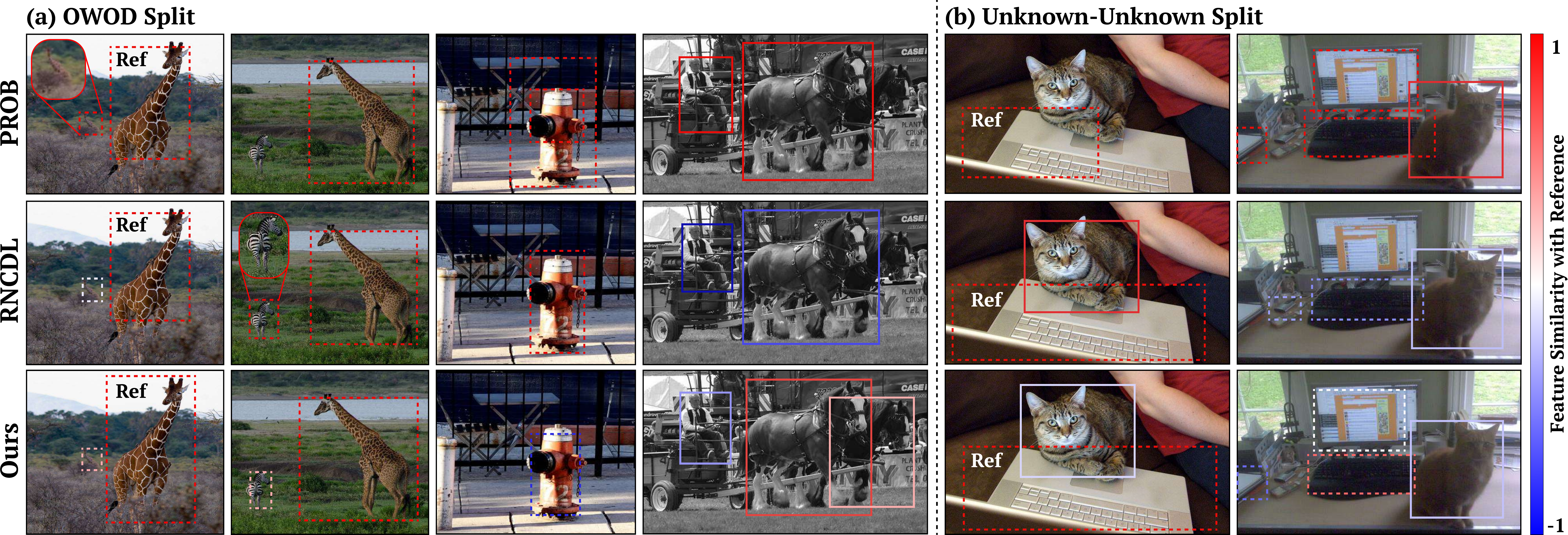}
\end{center}
\vspace{-0.2in}
\caption{
\textbf{Qualitative results for inter-proposal relationships on OWOD and Unknown-Unknown split.} Proposals with high feature similarity to the reference are shown red, while dissimilar proposals are in blue. \textit{Ours} successfully captures semantic similarities between both the known and unknown objects. For example, the reference \textit{giraffe} is similar to both an unknown \textit{giraffe} and a known \textit{horse}, while the \textit{fire hydrant} is highly dissimilar. In contrast, \textit{PROB} treats all proposals as highly similar. \textit{RNCDL}, despite using self-supervision to learn features, fails to capture meaningful semantics, mistakenly considering the \textit{giraffe} and \textit{fire hydrant} highly similar.
}
\label{fig:similarity_results}
\vspace{-0.1in}
\end{figure*}

\subsection{Embedding Transfer Module}
\hspace{0.15in} 
Previous self-supervised methods for learning semantically rich instance embeddings often suffer from noisy or weak supervision, highlighting the need for a robust and explicit supervisory signal to guide the detector's instance embeddings. Moreover, detectors must be able to appropriately embed not only \textit{known-unknown} objects but also \textit{unknown-unknown} objects, which requires a semantically rich feature space that accurately captures instance-wise relationships. To achieve this, we transfer the rich and generalizable features of the VFM to the detector. Specifically, pairwise similarities obtained from the VFM are used as relaxed labels for the contrastive loss, guiding the detector's instance embeddings to capture richer semantic relationships beyond the binary labels. Consequently, the detector’s instance embeddings are placed at varying distances from each other based on their semantic similarity.

We utilize the rich feature representations from DINOv2~\cite{oquab2023dinov2} as the source embedding. To distill this embedding space, a set of bounding boxes $\mathbf{B}=\{\mathbf{b}_i\}_{i=1}^{N}$ is utilized, comprising both known proposals obtained via Hungarian matching and refined unknown boxes from Sec.~\ref{sec:URM}, where $N$ denotes the total number of such boxes. First, the DINOv2 embedding $\mathbf{F} \in \mathbb{R}^D$ is extracted from the input image and resized to its original size. Proposal-wise source embeddings $\mathbf{S} = \{\mathbf{s}_i\}_{i=1}^{N}$, each with dimension $D$, are then computed by applying average pooling to the resized DINOv2 feature within each bounding box. Subsequently, the detector's proposal-wise instance embeddings $\mathbf{Z}=\{\mathbf{z}_i\}_{i=1}^{N}$, each with dimension $d$, are obtained by passing the query embeddings corresponding to $\mathbf{B}$ through an MLP head.

To distill the inter-sample relations of source embeddings to the detector's embedding, the semantic similarity of pairwise source embeddings is computed from their Euclidean distance, using a Gaussian kernel~\cite{kim2021embedding} as follows:
\begin{equation}
\label{eq:ETLR_weight}
\mathbf{w}_{ij} = 
    \exp\left({-\frac{\lVert \mathbf{s}_i - \mathbf{s}_j\rVert_2^2}{\sigma}}\right),
\end{equation}
where $\sigma$ is the kernel bandwidth. The instance embeddings are trained using the following relaxed contrastive loss~\cite{kim2021embedding}:
\begin{equation}
\label{eq:ETLR_loss}
\mathcal{L}_{et} =
\frac{1}{N}\sum_{i=1}^N
\sum_{j=1}^N
\left[
\mathbf{w}_{ij}\mathbf{d}_{ij}^2 +(1-\mathbf{w}_{ij})
\left[\delta - \mathbf{d}_{ij}
\right]_{+}^2
\right],
\end{equation}
where $\mathbf{d}_{ij}=\lVert\mathbf{z}_i - \mathbf{z}_j\rVert_{2}$ denotes the Euclidean distance between instance embeddings and $\delta$ is the margin.

Unlike binary contrastive loss, which focuses only on separating samples without considering their degree of similarity, Eq.~\ref{eq:ETLR_loss} adjusts the strength of pushing and pulling between instance embeddings based on semantic similarity derived from DINOv2. By incorporating these fine-grained inter-proposal relations during training, the generalizability of the detector's instance embedding space can be enhanced. The final loss is given by:
\begin{equation}
    \mathcal{L}=
    \mathcal{L}_{b}+
    \mathcal{L}_{c}+
    \mathcal{L}_{o}+
    \alpha\mathcal{L}_{b,\text{unk}}+
    \beta\mathcal{L}_{et}
\end{equation}
where $\alpha$ and $\beta$ denotes the weight coefficients.


\section{Experiments}
\label{sec:Experiments}

For more comprehensive understanding, please refer to the supplementary material (Sections~\ref{sec:Prob_Set}, \ref{sec:Add_Exp}, \ref{sec:Sup_Abl}, \ref{sec:Sup_Quanti} , and~\ref{sec:Sup_Quali}).

\subsection{Datasets}
\hspace{0.15in} 
We conducted experiments using splits of the MS-COCO~\cite{lin2014microsoft} and Pascal VOC~\cite{everingham2010pascal} datasets. In the \textit{OWOD split}~\cite{joseph2021towards}, all 80 classes are divided into four sequential tasks, $\{T_1,T_2,T_3,T_4\}$,  to simulate an incremental learning setting where each task contains 20 classes. During training for task $T_c$, only the 20 newly introduced classes in $T_c$ are labeled. Trained on this dataset, the detector is required to detect both the known classes from previously encountered tasks $\{T_t \mid t \le c\}$ and the unknown classes from future tasks $\{T_t \mid t > c\}$. Additionally, a few exemplars from previously known classes are stored in a replay buffer to mitigate catastrophic forgetting, which are used to fine-tune the detector at the end of each task~\cite{zohar2023prob}. By repeating this process, the detector continuously discovers and learns new object classes, gradually adapting to the open world.

The OWOD split allows unknown classes to appear in training images as unlabeled objects. However, it does not account for real-world scenarios where entirely novel objects that were never seen during training may appear during deployment. To better reflect these realistic conditions and to evaluate the detector's ability to detect and represent such \textit{unknown-unknown} objects, we propose a new Unknown-Unknown split. This split builds upon the OWDETR split~\cite{gupta2022ow}, which strictly separates superclasses across tasks. In addition, we exclude all images containing unknown classes from the training set, ensuring the detector has no prior exposure to these objects. This exclusion enables a more rigorous evaluation of the detector's ability to learn a generalizable feature embedding space for completely novel objects.

\begin{table*}[htbp]
\renewcommand\arraystretch{1.25}
\centering
\vspace{-0.1in}
\caption{\textbf{Quantitative results on Unknown-Unknown split.}  
We compared our method with the other approaches that simultaneously detect unknown objects and learn instance embeddings. While existing methods show decreased performance as \textit{unknown-unknown}s are never encountered during training, our method shows large improvement in both unknown object detection and feature embedding quality. This demonstrates that our method learns robust and generalizable instance embedding space.
} 
\label{tab:unkunk_results}
\resizebox{\textwidth}{!}{
\scriptsize
\begin{tabular}{l| ccc| ccc |ccc | c}\toprule
\multicolumn{1}{c|}{\textbf{Task IDs $\rightarrow$}}& \multicolumn{3}{c|}{\textbf{Task 1}}& \multicolumn{3}{c|}{\textbf{Task 2}} & \multicolumn{3}{c|}{\textbf{Task 3}}  & \multicolumn{1}{c}{\textbf{Task 4}} \\
\midrule
\multicolumn{1}{c|}{}   & \multicolumn{1}{c}{ \cellcolor[HTML]{fcfce0}Unknown}  & \cellcolor[HTML]{f4cccc}{Recall} & \cellcolor[HTML]{e4e5fa}Known &   \multicolumn{1}{c}{ \cellcolor[HTML]{fcfce0}Unknown} & \multicolumn{1}{c}{\cellcolor[HTML]{f4cccc}{Recall}} & \multicolumn{1}{c|}{\cellcolor[HTML]{e4e5fa}Known} &  \multicolumn{1}{c}{ \cellcolor[HTML]{fcfce0}Unknown} & \multicolumn{1}{c}{\cellcolor[HTML]{f4cccc}{Recall}} & \multicolumn{1}{c|}{\cellcolor[HTML]{e4e5fa}Known}   &\multicolumn{1}{c}{\cellcolor[HTML]{e4e5fa}Known}  \\
\multicolumn{1}{c|}{\multirow{-2}{*}{Metrics $\rightarrow$}}   & \cellcolor[HTML]{fcfce0} Recall & \cellcolor[HTML]{f4cccc} @1 &  \cellcolor[HTML]{e4e5fa}mAP & \cellcolor[HTML]{fcfce0}Recall& \cellcolor[HTML]{f4cccc} @1 & \multicolumn{1}{c|}{\cellcolor[HTML]{e4e5fa}mAP}& \cellcolor[HTML]{fcfce0}Recall  &  \cellcolor[HTML]{f4cccc} @1 &\multicolumn{1}{c|}{\cellcolor[HTML]{e4e5fa}mAP}    & \multicolumn{1}{c}{\cellcolor[HTML]{e4e5fa}mAP}\\
\midrule

\textit{PROB}\cite{zohar2023prob} & 21.04 & 3.32 & \textbf{67.65} & 29.85 & 7.81 & \textbf{44.80} & 33.37 & 12.62 & \textbf{39.65} & \textbf{38.70}  \\
\textit{OSODD}\cite{zheng2022towards} & 21.04 & 3.78 & \textbf{67.65} & 29.85 & 8.53 & \textbf{44.80} & 33.37 & 11.56 & \textbf{39.65} & \textbf{38.70}  \\
\textit{UC-OWOD}\cite{wu2022uc} & 14.80 & 2.41 & 28.52 & 19.47 & 4.87 & 25.04 & 21.98 & 7.43 & 26.96 & 38.07 \\
\textit{RNCDL}\cite{fomenko2022learning} & 21.91 & 1.72 & 66.73 & 27.81 & 2.40 & 42.51 & 32.17 & 4.96 & 39.17 & 37.08 \\

 \multirow{1}{*}{\textbf{\textit{Ours}}}   &\textbf{37.27}  &\textbf{11.21} & 66.82 &  \textbf{45.93}& \textbf{17.99} & 43.42  &\textbf{45.81} & \textbf{20.65} & 38.57  & 38.12 \\ 

\bottomrule 
\end{tabular}}
\vspace{-0.10in}
\end{table*}

\subsection{Experimental Setup}
\noindent\textbf{Implementation Details.}
We utilized \textit{PROB}~\cite{zohar2023prob} for the open-world object detector, which employs deformable DETR~\cite{zhu2020deformable} with a DINO-pretrained~\cite{caron2021emerging} Resnet-50 FPN backbone~\cite{gupta2022ow}. The model uses $100$ input queries, each with an embedding dimension of ${d=256}$. Training runs for 41 epochs, with the losses $\mathcal{L}_{b,\text{unk}}$ and $\mathcal{L}_{et}$ applied after the initial warmup period, using weight coefficients of ${\alpha=0.1}$, ${\beta=1.0}$. The batch size is set to 32, and the AdamW optimizer is employed with a learning rate of $3e^{-4}$.

For unknown box refinement, we employed the ViT-H SAM model~\cite{kirillov2023segment} to generate masks, using the top ${k=10}$ proposals with the highest objectness scores as prompts. For embedding transfer, we used the ViT-L DINOv2 model~\cite{oquab2023dinov2} with a feature dimension of ${D=1024}$. The remaining hyperparameters are set as follows: ${\kappa=0.5}$, ${\sigma=1.0}$, ${\delta=1.0}$.

\noindent\textbf{Evaluation Metrics.}
For known classes, the mean Average Precision~(mAP) is used for evaluation. However, for unknown objects, false positives become unreliable, as not every object in the dataset is labeled. Therefore, following previous works~\cite{liu2022opening, gupta2022ow, zohar2023prob}, the Unknown-Recall~(U-Recall) is used as a metric for unknown object detection.

To evaluate the quality of object embeddings, we first assign each ground truth object to the instance embedding of the predicted proposal with the highest overlap. Using these ground truth labels and corresponding features, we compute Recall@$K$ as a performance metric~\cite{kim2021embedding}, which measures how many objects have at least one sample from the same class among their $K$-nearest neighbors in the learned embedding space. When computing Recall@$K$, we assume that undetected ground truth objects have no valid corresponding samples in their $K$-nearest neighbors, as the detector failed to generate embeddings for them.

\noindent\textbf{Comparison Methods.}
We compared our method with several other approaches to demonstrate its effectiveness in detecting and extracting instance embeddings for unknown objects. For the base model, \textit{PROB}~\cite{zohar2023prob}, the detector's query embeddings are directly used as instance embeddings. We also evaluated methods that enhance feature quality through self-supervision. \textit{OSODD}~\cite{zheng2022towards} augments cropped proposals from the pretrained frozen detector and trains a separate proposal embedding network using contrastive learning~\cite{he2020momentum}. In contrast, \textit{UC-OWOD}~\cite{wu2022uc} and \textit{RNCDL}~\cite{fomenko2022learning} directly optimize the detector's instance embeddings. \textit{UC-OWOD} applies contrastive learning between instance embeddings of high-confidence unknown proposals, using only highly similar or highly dissimilar pairs as positive or negative samples. Meanwhile, \textit{RNCDL} clusters all instance embeddings based on similarity and treats all embeddings within the same cluster as positive pairs. All methods are reimplemented based on the \textit{PROB} framework.

\subsection{Experimental Results}
\begin{table*}[htbp]
\renewcommand\arraystretch{1.25}
\centering
\caption{
\textbf{Ablation results for each module.} Incorporating the \textit{Unknown Box Refine Module}~(URM) improves unknown detection performance~(Unknown Recall), while the \textit{Embedding Transfer Module}~(ETM) enhances the quality of instance embeddings~(Recall@1). Combining both modules achieves the highest feature embedding quality, with comparable unknown detection performance.
}
\label{tab:ablation_comp}
\vspace{-0.05in}
\resizebox{\textwidth}{!}{
\scriptsize
\begin{tabular}{l| ccc| ccc |ccc | c}\toprule
\multicolumn{1}{c|}{\textbf{Task IDs $\rightarrow$}}& \multicolumn{3}{c|}{\textbf{Task 1}}& \multicolumn{3}{c|}{\textbf{Task 2}} & \multicolumn{3}{c|}{\textbf{Task 3}}  & \multicolumn{1}{c}{\textbf{Task 4}} \\
\midrule
\multicolumn{1}{c|}{}   & \multicolumn{1}{c}{ \cellcolor[HTML]{fcfce0}Unknown}  & \cellcolor[HTML]{f4cccc}{Recall} & \cellcolor[HTML]{e4e5fa}Known &   \multicolumn{1}{c}{ \cellcolor[HTML]{fcfce0}Unknown} & \multicolumn{1}{c}{\cellcolor[HTML]{f4cccc}{Recall}} & \multicolumn{1}{c|}{\cellcolor[HTML]{e4e5fa}Known} &  \multicolumn{1}{c}{ \cellcolor[HTML]{fcfce0}Unknown} & \multicolumn{1}{c}{\cellcolor[HTML]{f4cccc}{Recall}} & \multicolumn{1}{c|}{\cellcolor[HTML]{e4e5fa}Known}   &\multicolumn{1}{c}{\cellcolor[HTML]{e4e5fa}Known}  \\
\multicolumn{1}{c|}{\multirow{-2}{*}{Metrics $\rightarrow$}}   & \cellcolor[HTML]{fcfce0} Recall & \cellcolor[HTML]{f4cccc} @1 &  \cellcolor[HTML]{e4e5fa}mAP & \cellcolor[HTML]{fcfce0}Recall& \cellcolor[HTML]{f4cccc} @1 & \multicolumn{1}{c|}{\cellcolor[HTML]{e4e5fa}mAP}& \cellcolor[HTML]{fcfce0}Recall  &  \cellcolor[HTML]{f4cccc} @1 &\multicolumn{1}{c|}{\cellcolor[HTML]{e4e5fa}mAP}    & \multicolumn{1}{c}{\cellcolor[HTML]{e4e5fa}mAP}\\

 \midrule

\textit{PROB}\cite{zohar2023prob} & 18.84 & 4.42 &  58.98 &  16.04 & 3.79 & 43.76 & 17.28 & 6.69 & 35.81 & \textbf{31.72}  \\

\textit{PROB}+URM & \textbf{30.79} & 5.53 & \textbf{59.23} & \textbf{27.29} & 5.40 & 44.16 & \textbf{29.10} & 9.29 & 35.85 & 31.23  \\
\textit{PROB}+ETM & 18.48 & 7.25 & 59.07 & 16.35 & 5.17 & 44.57 & 17.72 & 7.53 & 35.66 & 31.21  \\

 \multirow{1}{*}{\textbf{\textit{Ours}}}   & 30.56    &\textbf{11.69} & 58.89 &  26.74 & \textbf{10.19} & \textbf{45.00} & 28.21 & \textbf{13.25} & \textbf{36.31} & 31.46\\ 
\bottomrule 
\end{tabular}}
\vspace{-0.10in}
\end{table*}

\hspace{0.15in} 
We first compare our method with existing approaches that jointly perform unknown object detection and instance feature learning. We then present ablation studies on each module and baseline detector to demonstrate the effectiveness and compatibility of our approach.

\noindent\textbf{Comparison under OWOD split.} The quantitative results are summarized in Table.~\ref{tab:owod_results}. By transferring rich and robust supervisory signals from VFMs, our method outperforms all other approaches in both feature embedding quality and unknown object detection, while maintaining the performance on known object detection. \textit{OSODD}, which applies self-supervised learning to proposals from a frozen detector, shows only minor improvement. This suggests that training a separate embedding network on numerous inaccurate proposals leads to suboptimal results. On the other hand, methods that apply self-supervision directly to the detector's features show a decrease in performance. In particular, the significant performance drop in \textit{RNCDL}'s feature embedding quality indicates that numerous false positives lead to incorrect cluster assignments during self-supervised learning, severely degrading feature quality.

\begin{figure}[t]
\begin{center}
\includegraphics[width=1.0\linewidth]{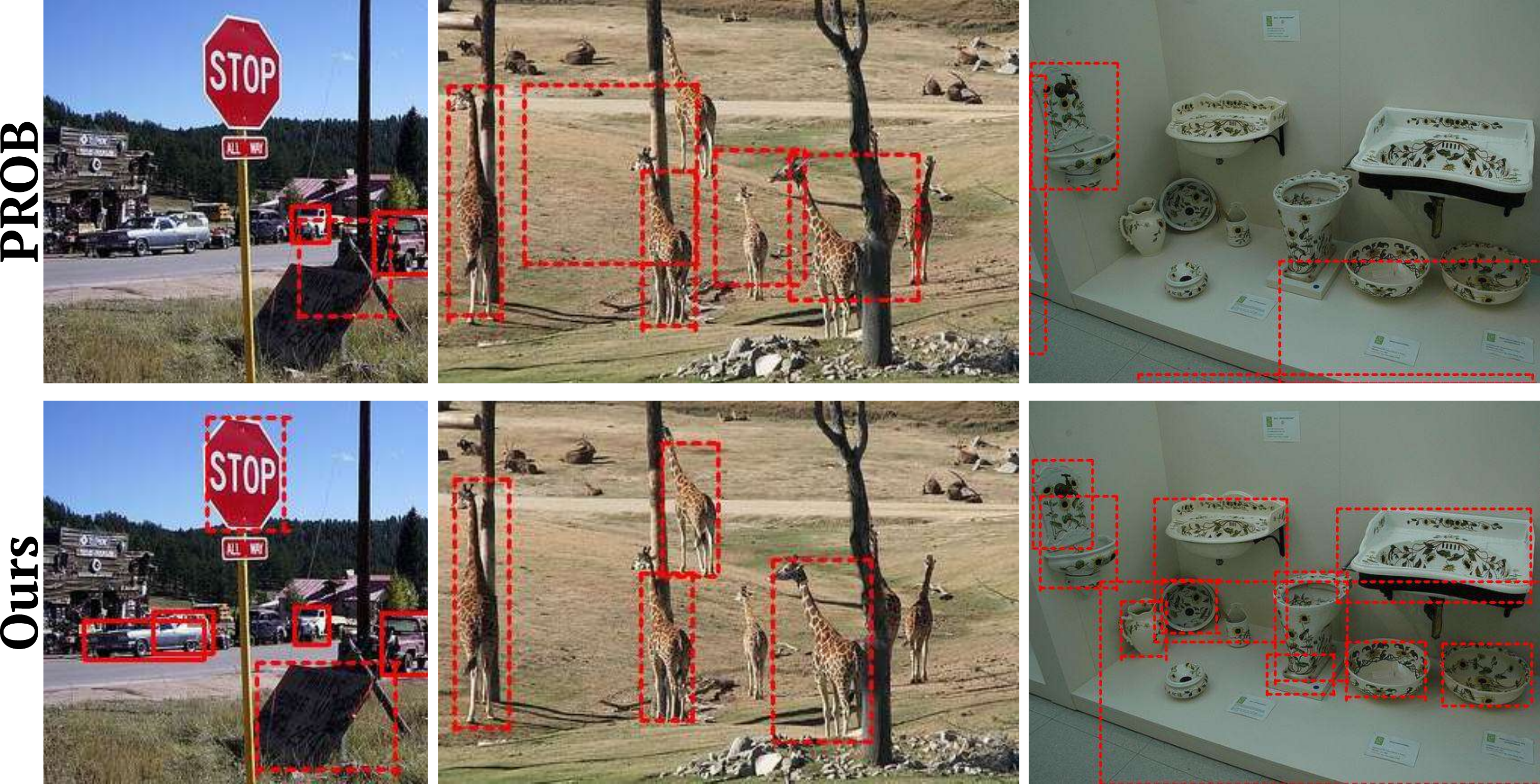}
\end{center}
\vspace{-0.15in}
\caption{
\textbf{Qualitative results of unknown object detection.} Unknown object detections from \textit{PROB}~(top row) and our model~(bottom row) are compared. By leveraging instance masks from SAM, our model achieves accurate localization.
}
\label{fig:detection_results}
\vspace{-0.20in}
\end{figure}

Despite the presence of numerous false positives and inaccurate boxes, our method successfully learns meaningful feature space and inter-proposal relations, as shown in Fig.~\ref{fig:similarity_results} (a). Using an embedding of unknown giraffe class as a reference, embeddings of other giraffes (unknown) and horses (known) show high similarity, while persons (known) and fire hydrants (unknown) appear clearly dissimilar. Moreover, Fig.~\ref{fig:detection_results} shows that our method detects unknown objects more precisely, demonstrating that the proposed unknown regression loss effectively enhances localization by leveraging SAM's instance masks.

\noindent\textbf{Comparison under Unknown-Unknown split.}
Table~\ref{tab:unkunk_results} shows that our method successfully embeds objects of \textit{unknown-unknown} classes, despite never observing them during training. In contrast, methods that rely on self-supervised learning of the detector's feature
still exhibit low performance.  This indicates that methods rigidly assigning positive or negative pairs during contrastive learning tend to learn a feature space tailored to discriminate the \textit{known-unknown} classes, reducing its generalizability. Similarly, \textit{OSODD} fails to form a sufficiently generalizable feature space due to numerous inaccurate proposals. Our method, as shown in Fig.~\ref{fig:similarity_results} (b), successfully captures the high similarity of keyboard embedding with other keyboards, and dissimilarity with cats, despite never observing keyboards during training. This demonstrates that our method learns a more generalizable feature space by distilling fine-grained instance-wise relationships from VFM.


\begin{table*}[htbp]
\renewcommand\arraystretch{1.25}
\centering
\vspace{-0.07in}
\caption{
\textbf{Ablation results for baseline Open-World Object Detector.} 
Applying our modules to other open-world object detectors, including \textit{OW-DETR}~\cite{gupta2022ow} and \textit{Hyp-OW}~\cite{doan_2024_HypOW}, our module improves Unknown Recall and Recall@1. This demonstrates that our method consistently enhances instance embedding quality and unknown object detection across different backbones. (see Sec~\ref{subsec:datasplit}. in the supplement)
}
\label{tab:ablation_back}
\resizebox{\textwidth}{!}{
\scriptsize
\begin{tabular}{l| ccc| ccc |ccc | c}\toprule
\multicolumn{1}{c|}{\textbf{Task IDs $\rightarrow$}}& \multicolumn{3}{c|}{\textbf{Task 1}}& \multicolumn{3}{c|}{\textbf{Task 2}} & \multicolumn{3}{c|}{\textbf{Task 3}}  & \multicolumn{1}{c}{\textbf{Task 4}} \\
\midrule
\multicolumn{1}{c|}{}   & \multicolumn{1}{c}{ \cellcolor[HTML]{fcfce0}Unknown}  & \cellcolor[HTML]{f4cccc}{Recall} & \cellcolor[HTML]{e4e5fa}Known &   \multicolumn{1}{c}{ \cellcolor[HTML]{fcfce0}Unknown} & \multicolumn{1}{c}{\cellcolor[HTML]{f4cccc}{Recall}} & \multicolumn{1}{c|}{\cellcolor[HTML]{e4e5fa}Known} &  \multicolumn{1}{c}{ \cellcolor[HTML]{fcfce0}Unknown} & \multicolumn{1}{c}{\cellcolor[HTML]{f4cccc}{Recall}} & \multicolumn{1}{c|}{\cellcolor[HTML]{e4e5fa}Known}   &\multicolumn{1}{c}{\cellcolor[HTML]{e4e5fa}Known}  \\
\multicolumn{1}{c|}{\multirow{-2}{*}{Metrics $\rightarrow$}}   & \cellcolor[HTML]{fcfce0} Recall & \cellcolor[HTML]{f4cccc} @1 &  \cellcolor[HTML]{e4e5fa}mAP & \cellcolor[HTML]{fcfce0}Recall& \cellcolor[HTML]{f4cccc} @1 & \multicolumn{1}{c|}{\cellcolor[HTML]{e4e5fa}mAP}& \cellcolor[HTML]{fcfce0}Recall  &  \cellcolor[HTML]{f4cccc} @1 &\multicolumn{1}{c|}{\cellcolor[HTML]{e4e5fa}mAP}    & \multicolumn{1}{c}{\cellcolor[HTML]{e4e5fa}mAP}\\

 \midrule

\textit{OW-DETR}\cite{gupta2022ow} & 9.84 & 2.26 &  57.02 &  6.26 & 1.50 & 39.31 & 5.90 & 1.66 & \textbf{28.90} & \textbf{24.62} \\

\textit{OW-DETR}+\textit{\textbf{Ours}} & \textbf{15.90} & \textbf{6.01} &  \textbf{57.18} &  \textbf{9.36} & \textbf{2.95} & \textbf{39.51} & \textbf{6.69} & \textbf{2.30} & 28.69 & 23.79  \\

\midrule
\textit{Hyp-OW}\cite{doan_2024_HypOW} & 22.64 & 6.93 &  58.77 & 20.13 & 5.60 & 39.21 &  24.05 & 10.14 & \textbf{31.86} & 29.20 \\

 \textit{Hyp-OW}+\textit{\textbf{Ours}}   & \textbf{26.21} & \textbf{10.28} & \textbf{58.83} & \textbf{27.71} & \textbf{10.92} & \textbf{39.42} & \textbf{30.59} & \textbf{15.20} & 30.28 & \textbf{29.84}\\

\bottomrule 
\end{tabular}}
\vspace{-0.2in}
\end{table*}

\noindent\textbf{Ablation Studies on Components.}
We conduct ablation studies to validate the effectiveness of each component, as summarized in Table~\ref{tab:ablation_comp}. The \textit{Embedding Transfer Module}~(\text{ETM}) improves the feature embedding quality as indicated by an increase in Recall@1, demonstrating that distilling VFM's instance-wise relationships enhances the detector's feature space. Additionally, the adoption of an \textit{Unknown Box Refine Module}~(\text{URM}) improves the accurate localization of boxes, increasing Unknown Recall. 
Overall, combining both components achieves the highest performance in terms of Recall@1 by learning the feature embedding from the accurate unknown boxes.

To evaluate the impact of our method on instance embedding quality, we visualize the instance embedding space using t-SNE, as shown in Fig.~\ref{fig:tsne}. Unlike \textit{PROB}, where embeddings are mixed and lack clear separation, our method produces a well-structured feature space with distinct class-wise clusters. Additionally, semantically similar animal classes appear closer together, while unrelated categories remain separate. This shows that our method effectively captures semantic relationships within the feature space.

\begin{figure}[h]
\begin{center}
\includegraphics[width=1.0\linewidth]{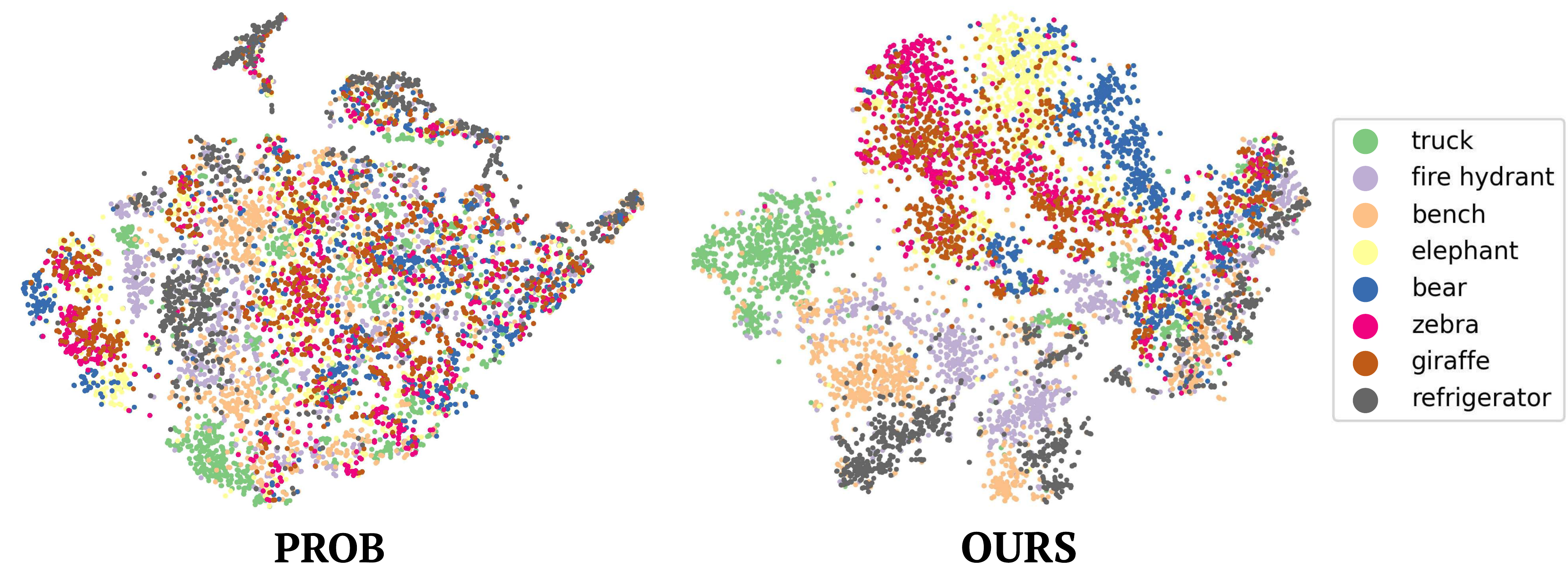}
\end{center}
\vspace{-0.1in}
\caption{
\textbf{t-SNE visualization of the learned instance embeddings.} Our method learns a rich instance embedding space, capturing semantic relationship between objects.
}
\label{fig:tsne}
\vspace{-0.1in}
\end{figure}

\noindent\textbf{Ablation Studies on Baseline Detector.}
Since our \textit{Unknown Box Refine Module} and \textit{Embedding Transfer Module} can be integrated into any open-world object detector, we conduct ablation studies on additional backbones beyond \textit{PROB}. As shown in Table~\ref{tab:ablation_back}, attaching our module to \textit{OW-DETR} and \textit{Hyp-OW} improves both unknown detection performance and instance-wise feature quality, as reflected in higher Unknown Recall and Recall@1. This demonstrates the effectiveness of our approach in enhancing open-world object detectors through instance feature learning.

\begin{table}[t]
\centering
\renewcommand{\arraystretch}{1.2}
\caption{
\textbf{Quantitative results for open-world tracking.} \textit{\textbf{Ours}} outperforms others in both unknown detection performance and the association of detections between frames. A-Accuracy represents the Association Accuracy~\cite{liu2022opening}.
}
\vspace{-0.05in}
\label{tab:tracking_result}
\resizebox{1.0\columnwidth}{!}{
    \tiny 
    \begin{tabular}{l|ccc}
        \toprule
        \multicolumn{1}{c|}{Metrics$\rightarrow$} & Unknown Recall & A-Accuracy & OWTA \\
        \midrule
         \textit{PROB~\cite{zohar2023prob}} & 48.67 & 8.49 & 19.72\\
         \textit{RNCDL~\cite{fomenko2022learning}} &  35.45 & 8.30 & 16.74 \\
         \textit{\textbf{Ours}} & \textbf{54.38} & \textbf{9.41} & \textbf{22.18}\\
        \bottomrule
    \end{tabular}
    }
    \vspace{-0.15in}
\end{table}

\subsection{Open-World Tracking}
\hspace{0.15in} 
To validate the applicability of our method, which detects both known and unknown objects while extracting semantically rich features, we apply it to the downstream task of open-world tracking~\cite{liu2022opening}. In tracking scenarios, objects often become occluded or undergo deformation, making robust features essential for accurate re-identification. By detecting and extracting semantically rich embeddings of both known and unknown objects, our method helps the tracker associate objects more reliably under open-world conditions. We adopt DeepSORT~\cite{wojke2017simple} as the tracker, which matches proposals across adjacent frames by combining the IoU of bounding boxes with the similarity between the detector's instance embeddings. The experiment uses a subset of the TAO-OW dataset~\cite{liu2022opening} containing objects with dynamic motion. Evaluation focuses only on unknown objects, following the OWTA metric~\cite{liu2022opening}, which combines unknown detection recall with association accuracy.

\begin{figure}[t]
\begin{center}
\includegraphics[width=1.0\linewidth]{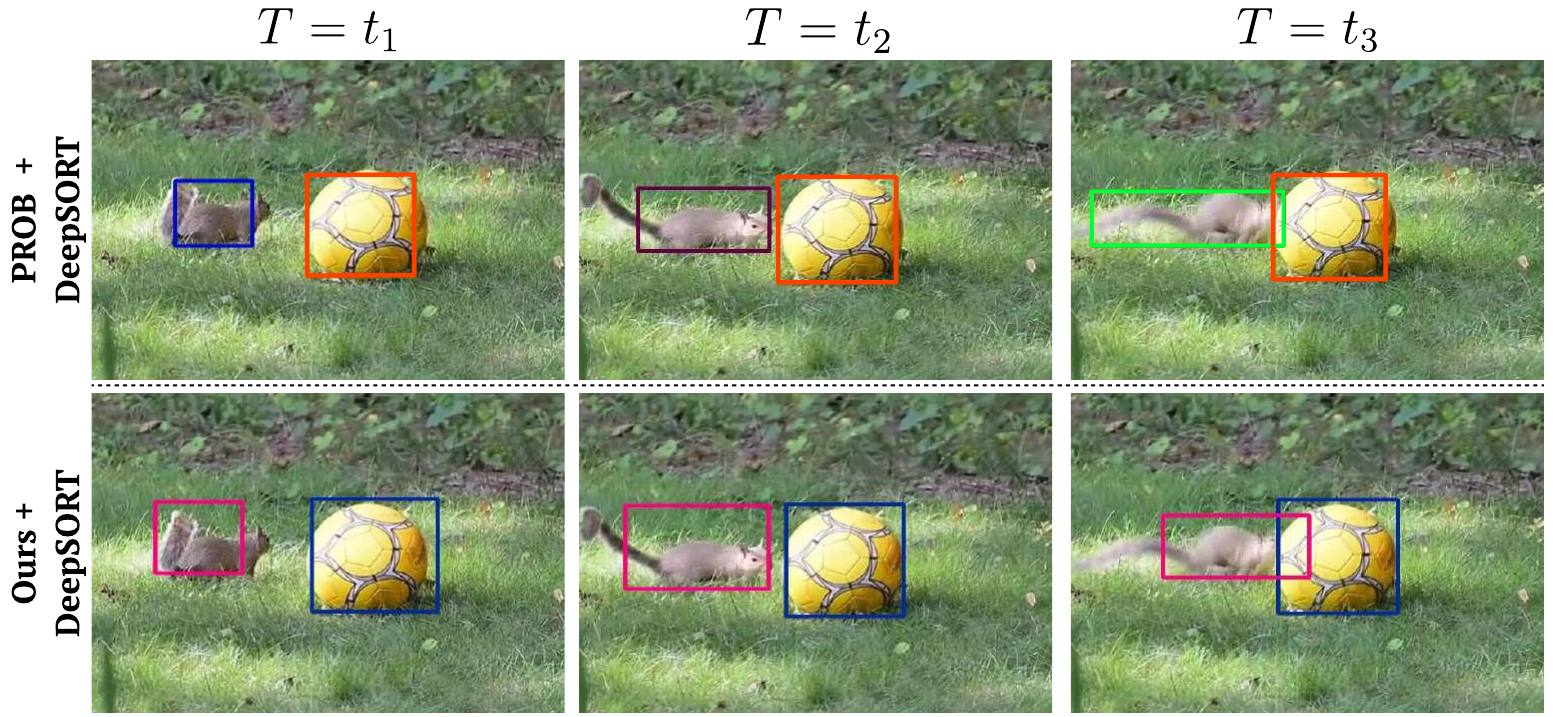}
\end{center}
\vspace{-0.15in}
\caption{
\textbf{Qualitative results for open-world tracking.} Bounding box colors represent track IDs, with both the \textit{squirrel} and \textit{ball} belonging to unknown classes. While the tracker using \textit{PROB} fails to correctly associate the \textit{squirrel} under shape changes, our method successfully tracks it by leveraging the semantically rich features extracted to associate proposals.
}
\label{fig:track}
\vspace{-0.15in}
\end{figure}

As shown in Table~\ref{tab:tracking_result}, our method improves both unknown object detection performance and association accuracy across frames, demonstrating the robustness of our proposals and instance embeddings. Fig.~\ref{fig:track} illustrates an example tracking scenario. When using outputs of \textit{PROB}, tracker fails to properly track the rapidly moving \textit{squirrel}, even though the object is successfully detected and meets the IoU threshold. This suggests that the instance embeddings from \textit{PROB} do not effectively capture the object's semantic characteristics, resulting in unreliable embeddings under deformation. In contrast, our method learns semantically rich instance embeddings, enabling the tracker to successfully re-identify the squirrel despite its motion and shape change.

\section{Conclusions}
\hspace{0.15in} This paper presents a method to extend open-world object detection by simultaneously detecting objects and extracting semantically rich features under open-world conditions. By distilling both predictions and features of Vision Foundation Models, the proposed method accurately localizes unknown objects and captures nuanced relationships between instances through instance embeddings. Extensive experiments demonstrate the effectiveness of our approach in learning a semantically rich and generalizable feature space for instance embeddings, expanding the applicability of open-world object detectors. To further enhance the real-world applicability, future work will explore online learning techniques to refine the feature space during inference.

\section{Acknowledgments}
This work was partly supported by Institute for Information \& communications Technology Technology Planning \& Evaluation(IITP) grant funded by the Korea government(MSIT)(RS-2019-II190075, Artificial Intelligence Graduate School Support Program(KAIST) and the Agency For Defense Development Grant funded by the Korean Government.

{
    \small
    \bibliographystyle{ieeenat_fullname}
    \bibliography{main}
}


\clearpage
\appendix
\etocdepthtag.toc{stoc} 
\etocsettagdepth{mtoc}{none} 
\etocsettagdepth{stoc}{subsection} 

\setcounter{page}{1}
\renewcommand{\thesection}{\Alph{section}}
\renewcommand{\thesubsection}{\thesection.\arabic{subsection}}
\renewcommand{\thefigure}{S\arabic{figure}}
\setcounter{figure}{0}  

\renewcommand{\thetable}{S\arabic{table}}
\setcounter{table}{0}

\renewcommand{\theequation}{S\arabic{equation}}
\setcounter{equation}{0}

\twocolumn[{%
\begin{center}
    \Large \textbf{Supplementary Materials} \\[1em]
\end{center}
}]
\tableofcontents


\section{Problem Setting: OWOD vs OVOD}
\label{sec:Prob_Set}
\begin{figure}[t]
\begin{center}
\includegraphics[width=1.0\linewidth]{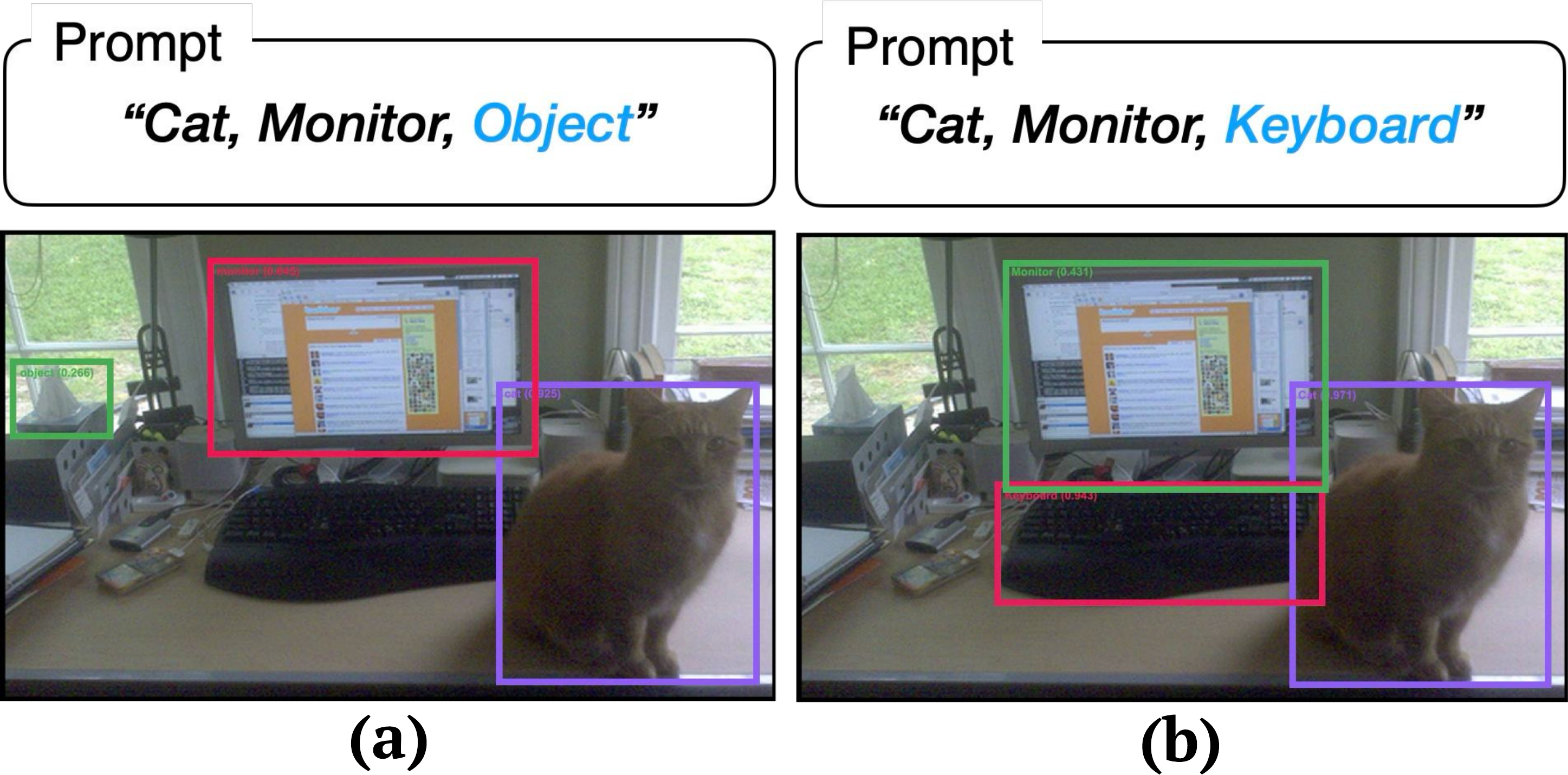}
\end{center}
\vspace{-0.1in}
\caption{
\textbf{Limitations of OVOD with text prompts.} Grounding DINO 1.6~\cite{ren2024grounding} is utilized for the experiment. (a) When the prompt is given as \textit{Cat, Monitor, Object}, the model detects the cat and the monitor, but fails to recognize the keyboard.
(b) When the prompt includes \textit{Keyboard}, the model successfully detects it. This demonstrates that OVOD methods are heavily dependent on explicit text prompts, and objects not included in the prompt cannot be detected.  
}
\label{fig:sup_dino}
\vspace{-0.15in}
\end{figure}

\begin{table*}[htbp]
\centering
\renewcommand{\arraystretch}{1.2}
\caption{
\textbf{Details for OWDETR split and Unknown-Unknown split.} Unknown-Unknown split is a subset of the OWDETR split and shares the same Semantic split. 
}
\label{subtab:data_split}
\resizebox{\textwidth}{!}{
    \begin{tabular}{l|cccc}
        \toprule
        \multicolumn{1}{c|}{\textbf{Task IDs}$\rightarrow$}  &\textbf{Task 1} & \textbf{Task 2} & \textbf{Task 3} & \textbf{Task 4} \\
        \midrule
         \multicolumn{1}{c|}{Semantic Split$\rightarrow$}  & Animals, Person, Vehicles & Appliances, Accessories, Outdoor, Furniture & Sports, Food & Electronic, Indoor, Kitchen\\
         \midrule
         OWDETR split Training images & 89490 & 55870 & 39402 & 38903 \\
         Unknown-Unknown split Training images & 26296 & 23509 & 28558 & 38903 \\
        \bottomrule
    \end{tabular}
    }
\end{table*}

\hspace{3pt} To clarify the scope of our work, we distinguish between the problem settings of Open Vocabulary Object Detection~(OVOD) and Open-World Object Detection~(OWOD), and highlight their complementary nature. OVOD focuses on detecting objects in an image based on given text prompts~\cite{gu2021open, kuo2022f, wu2023aligning, li2023distilling,wang2023open,wang2023learning,zang2022open,li2022grounded,liu2024grounding,ren2024grounding}. These methods typically leverage pretrained language models to fuse vision and language modalities, grounding the detector through explicit textual supervision. In contrast, OWOD, including our proposed approach, operates without any external input beyond the image itself. The goal is to detect both known objects (seen during training) and unknown objects (unseen categories), which can then be relabeled by human annotators and incrementally incorporated into the model to expand its knowledge. 

While OVOD naturally learns semantics of each object during grounding, previous OWOD methods focus on detecting unknown objects, without explicitly learning their semantics.
Some OVOD models like Grounding DINO~\cite{liu2024grounding} and GLIP~\cite{li2022grounded} are trained on vast, diverse detection datasets far exceeding COCO, with the primary objective of enabling zero-shot detection through text prompts. Consequently, their framework does not inherently address scenarios that lack text prompts or require incremental learning, which are central to the OWOD paradigm.

Our setting assumes that the set of objects to be detected is not known in advance and that corresponding text prompts are unavailable. This reflects a realistic open-world scenario where detectors inevitably encounter unexpected objects. In such cases, requiring text prompts for every possible object is impractical. Consequently, the fundamental difference between OWOD and OVOD lies in whether object detection depends on pre-specified text prompts. 

Due to this difference in formulation, we do not directly compare our method with OVOD approaches. OVOD is suitable for cases where explicit prompts are provided, whereas OWOD addresses scenarios where such prompts cannot be assumed. For example, as shown in Fig~\ref{fig:sup_dino}, Grounding DINO can detect a keyboard only when given the corresponding prompt. In deployment contexts such as autonomous driving, detectors must recognize previously unseen objects without relying on textual input—conditions that naturally fall under the OWOD setting rather than OVOD.

Moreover, a central objective of OWOD is not only to detect unknown objects but also to distinguish them from known classes and enable incremental learning, where newly identified categories can be incorporated into the model over time. These differences emphasize that OWOD and OVOD are complementary but fundamentally distinct research directions, rather than directly comparable approaches.

\section{Additional Experiments Material}
\label{sec:Add_Exp}
\subsection{Proposed Data Split}
\label{subsec:datasplit}
\hspace{0.15in} During real world deployment, a detector may encounter \textit{unknown-unknown} objects that were neither labeled nor present in the training data. Therefore, an open-world object detector must be capable of detecting and obtaining instance embeddings for never-encountered classes. To facilitate this evaluation, we introduce a novel data split for open-world object detection named the Unknown-Unknown split.

The \textit{Unknown-Unknown split} is built upon the OWDETR split, which was designed to minimize the presence of visually similar objects within the same data split. While the OWDETR split includes unknown class objects in the training set without labeling them, our proposed split goes a step further by entirely removing images containing unknown class objects from the training data. The details presented in Table ~\ref{subtab:data_split}.
To illustrate, consider the Task 1 setting of the OWDETR split, where the model is trained on known classes such as animals, persons, and vehicles. In this setup, the training set often includes images where known objects co-occur with instances of unknown, future-task classes—for example, a person in a kitchen with a donut or a sink. The standard protocol discards only the annotations for donut and sink but still uses the image for training. In contrast, our \textit{Unknown-Unknown split} excludes the entire image from the training data, thereby preventing the model from being visually exposed to any unknown objects.
Consequently, the model has no prior exposure to these \textit{unknown-unknown} objects. This data split enables a more rigorous evaluation of whether an open-world object detector can effectively identify truly unknown objects and capture their inter-proposal relationships by learning semantically rich feature representations.

\begin{figure}[t]
\begin{center}
\includegraphics[width=1.0\linewidth]{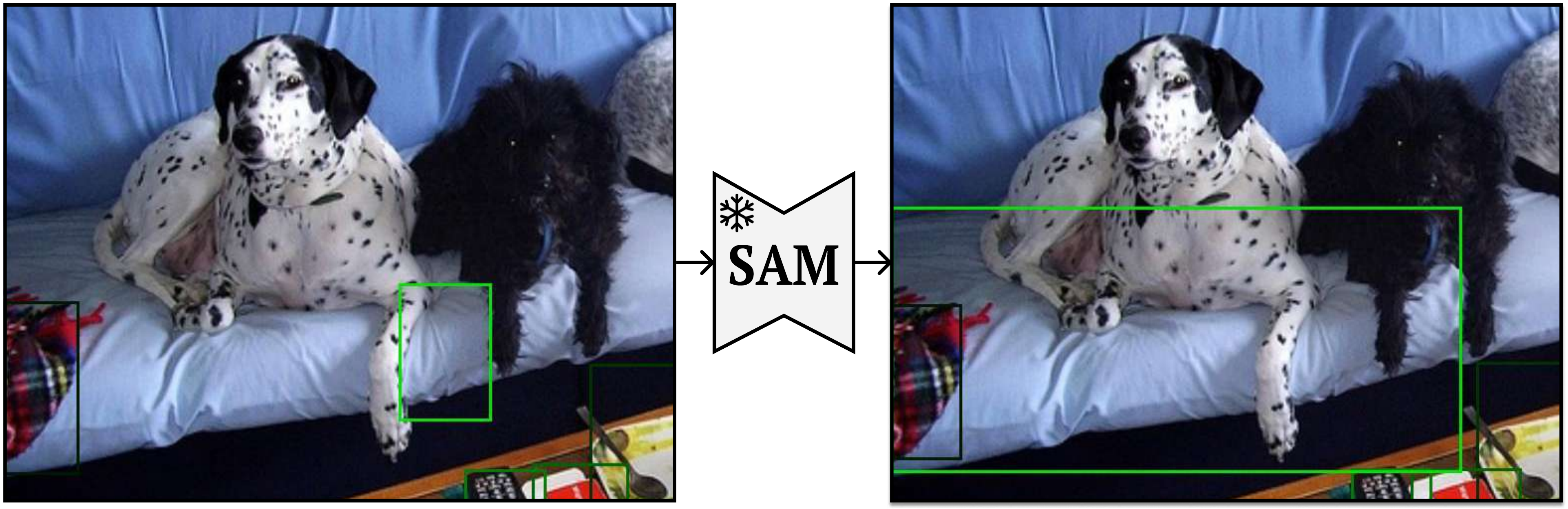}
\end{center}
\vspace{-0.1in}
\caption{
\textbf{A failure case of the SAM-based box refinement.} \textbf{Left:} The initial bounding boxes predicted by the detector. \textbf{Right:} The refined boxes generated from SAM's masks. Boxes of the same color correspond to the proposals before and after refinement. The green box highlights a significant discrepancy that can occur without IoU thresholding, underscoring its importance.
}
\label{fig:sam_failure}
\vspace{-0.15in}
\end{figure}

\subsection{Motivation for Evaluation Metrics}
\hspace{0.15in} 
To evaluate the feature embedding quality of open-world detectors, existing methods typically cluster detected unknown objects into a predefined number of unknown classes, assign ground-truth labels to each cluster, and assess performance using metrics such as mAP. However, this approach assumes a fixed number of unknown classes and clusters all detected objects accordingly. In practice, not all objects in a dataset are labeled, and open-world object detectors may identify unlabeled objects, making clustering unreliable. Furthermore, numerous inaccurate detections further degrade clustering quality, rendering the evaluation metric unreliable.

To this end, we replace discrete clustering-based evaluation with a direct assessment of feature space quality using Recall@K. By employing Recall@K as an evaluation metric for feature learning, we can evaluate whether instances of the same unknown class are embedded closely without relying on a predefined number of unknown classes.

\subsection{Additional Implementation Details}
\hspace{0.15in} During the initial training phase, only \(L_o\), \(L_c\), and \(L_b\) are optimized. The \textit{Embedding Transfer Module} and \textit{Unknown Box Refine Module} are introduced after 14 and 35 epochs, respectively. For the \textit{Unknown Box Refine Module}, four regular grid points are sampled from each input unknown proposal, and used as prompts to generate a instance mask. The refined unknowns are defined as the smallest bounding box containing this mask. Bounding boxes are used for training only if their IoU before and after refinement is 0.5 or higher, minimizing performance degradation on known classes. This ensures the detector from training with wrong supervisions, such as depicted in Fig~\ref{fig:sam_failure}

\section{Ablation and Component Analysis}
\label{sec:Sup_Abl}
\subsection{Additional Ablation for Baseline Detector.}
\hspace{0.15in}
Our approach is designed as modular components that can be integrated into any baseline open-world object detector. We primarily adopt \textit{PROB} as the baseline detector to highlight the effectiveness of our method in improving both unknown object detection and instance-level feature representation. Additionally, we validate its general applicability by conducting experiments with multiple OWOD baselines Table 4 and Table~\ref{rebtab:compare}.

\begin{table}[t]
\centering
\renewcommand{\arraystretch}{1.25}
\setlength{\abovecaptionskip}{4pt}
\caption{
 \textbf{Ablation results for base Open World Object Detector.} U-Recall denotes the unknown detection recall. Incorporating our module enhances both instance embedding quality and unknown detection performance.
}
\label{rebtab:compare}
\resizebox{1.0\columnwidth}{!}{
\tiny
    \begin{tabular}{l|ccc}
        \toprule
\multicolumn{1}{c|}{\textbf{Task IDs}$\rightarrow$} & \multicolumn{3}{c}{\textbf{Task 1}}\\
\midrule
\multicolumn{1}{c|}{Metrics$\rightarrow$} & \cellcolor[HTML]{fcfce0}{U-Recall} & \cellcolor[HTML]{f4cccc}{Recall@1} & \cellcolor[HTML]{e4e5fa}{Known mAP}\\
\midrule
\textit{CAT}~\cite{ma2023cat}                    & 9.06 & 3.07 & 59.33 \\
\textit{CAT}+\textbf{\textit{Ours}}  & \textbf{11.67} & \textbf{4.81} & \textbf{59.75} \\
\midrule
\textit{OrthoDet}~\cite{sun2024exploring}                    & 21.56 & 11.84 & \textbf{60.83} \\
\textit{OrthoDet}+\textbf{\textit{Ours}}  & \textbf{29.74} & \textbf{13.35} & 59.41 \\
\bottomrule
    \end{tabular}
    }
\end{table}

\subsection{Comparison with using SAM and DINOv2 features for ETM}
\hspace{0.15in}
Table~\ref{rebtab:ablation_vfm} presents results of distilling feature similarity from SAM and DINOv2. Although embedding transfer can use any source model, we use DINOv2 for its superior feature learning performance.

\begin{table}[t]
\centering
\renewcommand{\arraystretch}{1.2}
\setlength{\abovecaptionskip}{4pt}
\caption{
 \textbf{Results using SAM and DINOv2 features for embedding transfer.} U-Recall denotes the unknown detection recall. Distilling DINOv2 features further improves feature quality.
}
\label{rebtab:ablation_vfm}
\resizebox{1.0\columnwidth}{!}{
\tiny
    \begin{tabular}{l|ccc}
        \toprule
\multicolumn{1}{c|}{Metrics$\rightarrow$} & \cellcolor[HTML]{fcfce0}{U-Recall} & \cellcolor[HTML]{f4cccc}{Recall@1} & \cellcolor[HTML]{e4e5fa}{Known mAP}\\
\midrule
\textbf{\textit{Ours}} \text{(w. SAM)}      & \textbf{30.63} & 10.38 & \textbf{59.38} \\
\textbf{\textit{Ours}} \text{(w. DINOv2)}   & 30.56 & \textbf{11.69} & 58.89 \\
\bottomrule
    \end{tabular}
    }
\end{table}

\subsection{Ablation Studies of Distillation Method}
\hspace{0.15in} To evaluate the effectiveness of the \textit{Embedding Transfer Module}, we compare our method with other distillation approaches using VFM embeddings. Specifically, we consider a baseline that directly aligns the detector's instance embeddings with source embeddings from DINOv2 using L1 loss~(L1 Distill). Since the embedding dimension $D$ of DINOv2 differs from the detector’s instance embedding dimension $d$, we adjust the output dimension of the MLP layer for instance embedding accordingly. Experiments are conducted on a detector trained with Task 1 of the OWOD split.

The results are shown in Table~\ref{suptab:distill}.
Our model demonstrates superior embedding quality compared to the L1 loss method, which simply forces detector features to match VFM features. This suggests that directly mimicking the high-dimensional feature space of VFMs is suboptimal. By using feature similarity as a weight in contrastive loss during distillation, our method learns a semantically rich feature space.

\begin{table}[t]
\centering
\renewcommand{\arraystretch}{1.3}
\caption{
 \textbf{Ablation results for the distillation method.} U-Recall denotes the unknown detection recall, and L1 Distill denotes the direct distillation of DINOv2 features via L1 loss. Our method outperforms the L1 Distill method and shows superior feature representation quality.
}
\label{suptab:distill}
\resizebox{1.0\columnwidth}{!}{
\tiny
    \begin{tabular}{l|ccc}
        \toprule
\multicolumn{1}{c|}{Metrics$\rightarrow$}                   & \cellcolor[HTML]{fcfce0}{U-Recall} & \cellcolor[HTML]{f4cccc}{Recall@1} & \cellcolor[HTML]{e4e5fa}{Known mAP}\\
\midrule
\textit{PROB}                    & 18.84 & 4.42 & 58.98\\
\textit{PROB} + L1 Distill  & 31.19 & 8.69 & 59.20 \\
Ours                    & \textbf{31.54} & \textbf{11.40} & \textbf{59.12} \\
        \bottomrule
    \end{tabular}
    }
\end{table}


\begin{table*}[t]
\renewcommand\arraystretch{1.25}
\centering
\caption{
\textbf{Ablation results for each component in Unknown-Unknown split.} Incorporating the \textit{Embedding Transfer Module}~(\textbf{ETM}) and \textit{Unknown Box Refine Module}~(\textbf{URM}) enhances the instance embedding quality and unknown detection performance, respectively. 
}
\label{tab:sup_ablation_comp}
\resizebox{\textwidth}{!}{
\tiny
\begin{tabular}{l| ccc| ccc |ccc | c}\toprule
\multicolumn{1}{c|}{\textbf{Task IDs $\rightarrow$}}& \multicolumn{3}{c|}{\textbf{Task 1}}& \multicolumn{3}{c|}{\textbf{Task 2}} & \multicolumn{3}{c|}{\textbf{Task 3}}  & \multicolumn{1}{c}{\textbf{Task 4}} \\\midrule
\multicolumn{1}{c|}{}   & \multicolumn{1}{c}{ \cellcolor[HTML]{fcfce0}Unknown}  & \cellcolor[HTML]{f4cccc}{Recall} & \cellcolor[HTML]{e4e5fa}Known &   \multicolumn{1}{c}{ \cellcolor[HTML]{fcfce0}Unknown} & \multicolumn{1}{c}{\cellcolor[HTML]{f4cccc}{Recall}} & \multicolumn{1}{c|}{\cellcolor[HTML]{e4e5fa}Known} &  \multicolumn{1}{c}{ \cellcolor[HTML]{fcfce0}Unknown} & \multicolumn{1}{c}{\cellcolor[HTML]{f4cccc}{Recall}} & \multicolumn{1}{c|}{\cellcolor[HTML]{e4e5fa}Known}   &\multicolumn{1}{c}{\cellcolor[HTML]{e4e5fa}Known}  \\

\multicolumn{1}{c|}{\multirow{-2}{*}{Metrics $\rightarrow$}}   & \cellcolor[HTML]{fcfce0} Recall & \cellcolor[HTML]{f4cccc} @1 &  \cellcolor[HTML]{e4e5fa}mAP & \cellcolor[HTML]{fcfce0}Recall& \cellcolor[HTML]{f4cccc} @1 & \multicolumn{1}{c|}{\cellcolor[HTML]{e4e5fa}mAP}& \cellcolor[HTML]{fcfce0}Recall  &  \cellcolor[HTML]{f4cccc} @1 &\multicolumn{1}{c|}{\cellcolor[HTML]{e4e5fa}mAP}    & \multicolumn{1}{c}{\cellcolor[HTML]{e4e5fa}mAP}\\

 \midrule

\textit{PROB} & 21.04 & 3.32 & 67.65 & 29.85 & 7.81 & \textbf{44.80} & 33.37 & 12.62 & \textbf{39.65} & 38.70  \\

\textit{PROB}+URM & 36.71 & 5.14 & 66.77 & 44.80 & 9.85 & 43.71 & 44.84 & 15.29 & 38.16 &  38.27 \\
\textit{PROB}+ETM & 21.83 & 5.49 & \textbf{67.68} & 31.80 & 11.15 & 43.78 & 33.61 & 13.96 & 39.58 & \textbf{38.73}  \\

 \multirow{1}{*}{\textbf{\textit{Ours}}}  & \textbf{37.27} & \textbf{11.21} & 66.82 & \textbf{45.93} & \textbf{17.99} & 43.42 & \textbf{45.81} & \textbf{20.65} & 38.57 & 38.12 \\ 
\bottomrule 
\end{tabular}}
\end{table*}

\subsection{Ablation Studies of Module under Unknown-Unknown split}
\hspace{0.15in} 
We conduct ablation studies under the \textit{Unknown-Unknown split} to validate the effectiveness of each component. The results are summarized in Table~\ref{tab:sup_ablation_comp}. Consistent with the findings from the OWOD split, integrating the \textit{Embedding Transfer Module} improves the quality of instance embeddings by distilling the rich semantic information of VFMs, while the \textit{Unknown Box Refine Module} enhances unknown object detection performance. When both modules are used together, feature learning is applied to refined proposals, leading to a significant improvement in feature embedding quality.


\subsection{Ablation Studies of each Module's Hyperparameters}
\hspace{0.15in} 
We performed a comprehensive hyperparameter ablation study to validate the robustness of the ETM and URM modules within the OWOD split. The experimental results are detailed in Table~\ref{tab:hyperparams_ablation}. Our model demonstrates insensitivity to variations in each hyperparameter. Furthermore, we observed that varying parameters like $k$ and $\kappa$ maintained performance without compromising the known mAP of the PROB baseline.

\begin{table}[t]
\centering
\renewcommand{\arraystretch}{1.3}
\caption{
    \textbf{Ablation results for each hyperparameters of ETM and URM on OWOD split.} The first row shows our default hyperparameter results. In subsequent rows, we vary a single hyperparameter, highlighted in green. 
}
\label{tab:hyperparams_ablation}
\resizebox{1.0\columnwidth}{!}{%
\scriptsize
    \begin{tabular}{c|c|c|c||ccc}
        \toprule
        \multicolumn{4}{c||}{\textbf{Task IDs}$\rightarrow$} & \multicolumn{3}{c}{\textbf{Task 1}}\\
        \midrule
        \multicolumn{2}{c|}{\textbf{ETM}} & \multicolumn{2}{c||}{\textbf{URM}} & \cellcolor[HTML]{fcfce0}Unknown &  \cellcolor[HTML]{f4cccc}Recall &  \cellcolor[HTML]{e4e5fa}Known \\
        \cmidrule(lr){1-1} \cmidrule(lr){2-2} \cmidrule(lr){3-3} \cmidrule(lr){4-4}
        \textbf{$\delta$} & \textbf{$\sigma$} & \textbf{$k$} & \textbf{$\kappa$} & \cellcolor[HTML]{fcfce0}Recall & \cellcolor[HTML]{f4cccc}@1 & \cellcolor[HTML]{e4e5fa}mAP \\
        \midrule
        \rowcolor[HTML]{EFEFEF}
        1 & 1 & 10 & 0.5 & 30.56 & 11.69 & 58.89 \\
        \midrule
        1 & 1 & 10 & \cellcolor[HTML]{d9ead3}0.45 & 31.42 & \textbf{11.98} & 58.38 \\
        1 & 1 & 10 & \cellcolor[HTML]{d9ead3}0.55 & 30.11 & 11.47 & 59.12 \\
        1 & 1 & \cellcolor[HTML]{d9ead3}5 & 0.5 & 28.51 & 10.58 & \textbf{59.31} \\
        1 & 1 & \cellcolor[HTML]{d9ead3}15 & 0.5 & \textbf{32.01} & 11.89 & 58.80 \\
        1 & \cellcolor[HTML]{d9ead3}0.5 & 10 & 0.5 & 31.55 & 10.69 & 59.02 \\
        1 & \cellcolor[HTML]{d9ead3}2.0 & 10 & 0.5 & 31.52 & 11.59 & 58.95 \\
        \cellcolor[HTML]{d9ead3}0.9 & 1 & 10 & 0.5 & 30.70 & 11.20 & 58.89 \\
        \cellcolor[HTML]{d9ead3}1.1 & 1 & 10 & 0.5 & 31.68 & 11.93 & 58.79 \\
        \bottomrule
    \end{tabular}%
}
\end{table}

\section{Additional Quantitative Results}
\label{sec:Sup_Quanti}
\subsection{Quantitative results on OWDETR split}
\hspace{0.15in}
Table~\ref{rebtab:sowodb} demonstrates that our method outperforms other models in both unknown object detection and feature learning on OWDETR split. This trend is consistent with the results on the Unknown-Unknown split shown in Table 2,3.
The Unknown-Unknown split is a subset of the OWDETR split as described in \ref{subsec:datasplit}, which explains the similarity in performance trends.

\begin{table}[h]
\centering
\renewcommand{\arraystretch}{1.2}
\setlength{\abovecaptionskip}{4pt}
\caption{
 \textbf{Results on OWDETR split.} U-Recall denotes the unknown detection recall. Our method outperforms others in unknown detection and feature quality.
}
\label{rebtab:sowodb}
\resizebox{1.0\columnwidth}{!}{
\tiny
    \begin{tabular}{l|ccc}
        \toprule
\multicolumn{1}{c|}{\textbf{Task IDs}$\rightarrow$} & \multicolumn{3}{c}{\textbf{Task 1}}\\
\midrule
\multicolumn{1}{c|}{Metrics$\rightarrow$} & \cellcolor[HTML]{fcfce0}{U-Recall} & \cellcolor[HTML]{f4cccc}{Recall@1} & \cellcolor[HTML]{e4e5fa}{Known mAP}\\
\midrule
\textit{PROB}    & 16.77 & 2.27 & \textbf{73.09} \\
\textit{OSODD}   & 16.77 & 2.22 & \textbf{73.09} \\
\textit{UC-OWOD} & 10.62 & 1.27 & 22.84\\
\textit{RNCDL}   & 17.09 & 1.52 & 72.34\\
\midrule
\textit{PROB}+URM    & \textbf{29.68} & 3.59 & 71.21 \\
\textit{PROB}+ETM    & 15.31 & 3.45 & 72.8 \\
\textbf{\textit{Ours}}    & 29.51 & \textbf{9.05} & 71.90 \\
\bottomrule
    \end{tabular}
    }
\end{table}

\subsection{Quantitative Results of Efficiency Comparisons including Direct VFM usage}
\hspace{0.15in}
Table~\ref{rebtab:infertime} reports runtime and memory costs, including a comparison with \textit{Ours}+DINO, which pools DINOv2 features from proposals at inference. Our method introduces minimal overhead relative to \textit{PROB}, requiring only a few additional MLP layers for instance feature extraction, and thus achieves comparable inference speed. In contrast, \textit{Ours}+DINO Pool increases computation. The number of trainable parameters grows by only 0.5\%, since SAM and DINO features are extracted once per image and reused during training.

We distill knowledge from VFMs rather than use them directly for two reasons: (1) VFMs are not designed for open-world detection, which requires identifying and incrementally learning unknown objects, and (2) direct use at inference is computationally costly.

\begin{table}[t]
\centering
\renewcommand{\arraystretch}{1.2}
\setlength{\abovecaptionskip}{4pt}
\small 
\caption{
    \textbf{Efficiency comparisons including direct VFM usage.} Our method enables effective feature learning with minimal computational overhead, compared to directly using VFM.
}

\label{rebtab:infertime}
 \resizebox{1.0\columnwidth}{!}{
  \scriptsize
    \begin{tabular}{l|ccc}
        \toprule
        \multicolumn{1}{c|}{\textbf{Metrics} $\downarrow$} & \textit{PROB} & \textit{Ours} & \textit{Ours}+DINO Pool \\
        \midrule
        Inference Time (msec) & 47.1 & 47.2 & 80.9 \\
        Inference \# Params & 39.7M & 39.9M & 344M \\
        Peak Inference Memory (MiB) & 3030 & 3036 & 3036 \\ 
        \midrule
        Train Time (relative) & x1.00 & x1.74 & x1.74 \\
        Peak Train Memory (MiB) & 80660 & 80700 & 80700 \\ 
        \midrule
        \rowcolor[HTML]{f4cccc} 
        Recall@1 & 4.42 & 11.69 & 15.53 \\
        \bottomrule
    \end{tabular}
    }
\end{table}

\subsection{Class Similarity Analysis}
\hspace{0.15in} We evaluate whether unknown objects are closely embedded to their semantically similar known classes in the feature space.
Specifically, we compute Recall@1 for unknown objects, which measures the proportion of unknown objects whose nearest known class shares the same superclass. The nearest known class is identified by comparing the unknown object's embedding to the centroids of known classes, where each centroid is the average embedding of all objects classified as that class.
The results presented in Table~\ref{tab:class_similarity_results} demonstrate that our method outperforms both the baseline and other self-supervised learning approaches. By leveraging the rich feature space of VFMs, our model effectively captures the relationships between known and unknown objects, embedding unknown objects closer to their semantically similar known counterparts.

\begin{table}[t]
\centering
\renewcommand{\arraystretch}{1.3}
\caption{
 \textbf{Class similarity analysis on animal and vehicle superclasses.} Our method embeds unknown objects closer to their semantically similar known classes, preserving meaningful relationships in the feature space.
}
\label{tab:class_similarity_results}
\resizebox{1.0\columnwidth}{!}{
    \begin{tabular}{l|cc|cc}
        \toprule
        \multicolumn{1}{c|}{\textbf{Superclass}$\rightarrow$} & \multicolumn{2}{c|}{\textbf{Animal}} & \multicolumn{2}{c}{\textbf{Vehicle}}\\
        \midrule
        \multicolumn{1}{c|}{Metrics$\rightarrow$} & \multicolumn{1}{c}{\cellcolor[HTML]{fcfce0}{Unknown Recall}} & \multicolumn{1}{c|}{\cellcolor[HTML]{f4cccc}{Recall@1}} & \multicolumn{1}{c}{\cellcolor[HTML]{fcfce0}{Unknown Recall}} & \multicolumn{1}{c}{\cellcolor[HTML]{f4cccc}{Recall@1}} \\
        \midrule
         \textit{PROB} & 87.80  & 43.30 & 60.97 & 33.12 \\
         \textit{OSODD} &87.80 &19.14 & 60.97 & 13.50 \\
         \textit{UC-OWOD} & 83.61 &45.22 & 59.07 & 41.35 \\
         \textit{RNCDL} &87.68 &8.13 &59.49 & 4.43 \\
         \textit{Ours} &\textbf{89.00} & \textbf{59.33}& \textbf{65.82} & \textbf{47.68}\\
        \bottomrule
    \end{tabular}
    }
\end{table}


\subsection{Quantitative results of Unknown Confusion Metrics}
\hspace{0.15in}
Our method slightly increases confusion metrics, possibly due to the refinement of misclassified unknown boxes. The results are shown in Table~\ref{rebtab:confusion}. Moreover, learning nuanced instance relationships rather than enforcing strict class boundaries may weaken the separation between known and unknown classes.


\section{Additional Qualitative Results}
\label{sec:Sup_Quali}

\begin{table}[t]
\centering
\renewcommand{\arraystretch}{1.3}
\setlength{\abovecaptionskip}{4pt}
\caption{
 \textbf{Results for unknown confusion metrics.} WI denotes Wilderness Impact, and A-OSE denotes Absolute Open-Set Error.
}
\label{rebtab:confusion}
\resizebox{1.0\columnwidth}{!}{
\scriptsize
    \begin{tabular}{l|cc|cc|cc}
        \toprule
        \multicolumn{1}{c|}{\textbf{Task IDs}$\rightarrow$} & \multicolumn{2}{c|}{\textbf{Task 1}} & \multicolumn{2}{c|}{\textbf{Task 2}} & \multicolumn{2}{c}{\textbf{Task 3}}\\
        \midrule
        \multicolumn{1}{c|}{Metrics$\rightarrow$} & WI & A-OSE & WI & A-OSE & WI & A-OSE \\
\midrule
\textit{PROB}           & 0.0550 & 4956 & 0.0317 & 5596 & 0.0165 & 3122  \\
\textit{PROB}+URM       & 0.0642 & 6821 & 0.0288 & 4399 & 0.0197 & 4360 \\
\textit{PROB}+ETM       & 0.0554 & 5065 & 0.0277 & 3667 & 0.0169 & 2808 \\
\textbf{\textit{Ours}}  & 0.0621 & 6056 & 0.0359 & 7739 & 0.0232 & 5099 \\
\bottomrule
    \end{tabular}
    }
\end{table}

\subsection{Qualitative Results of Inter-proposal Relationships.}
\begin{figure*}[t]
\begin{center}
\includegraphics[width=1.0\linewidth]{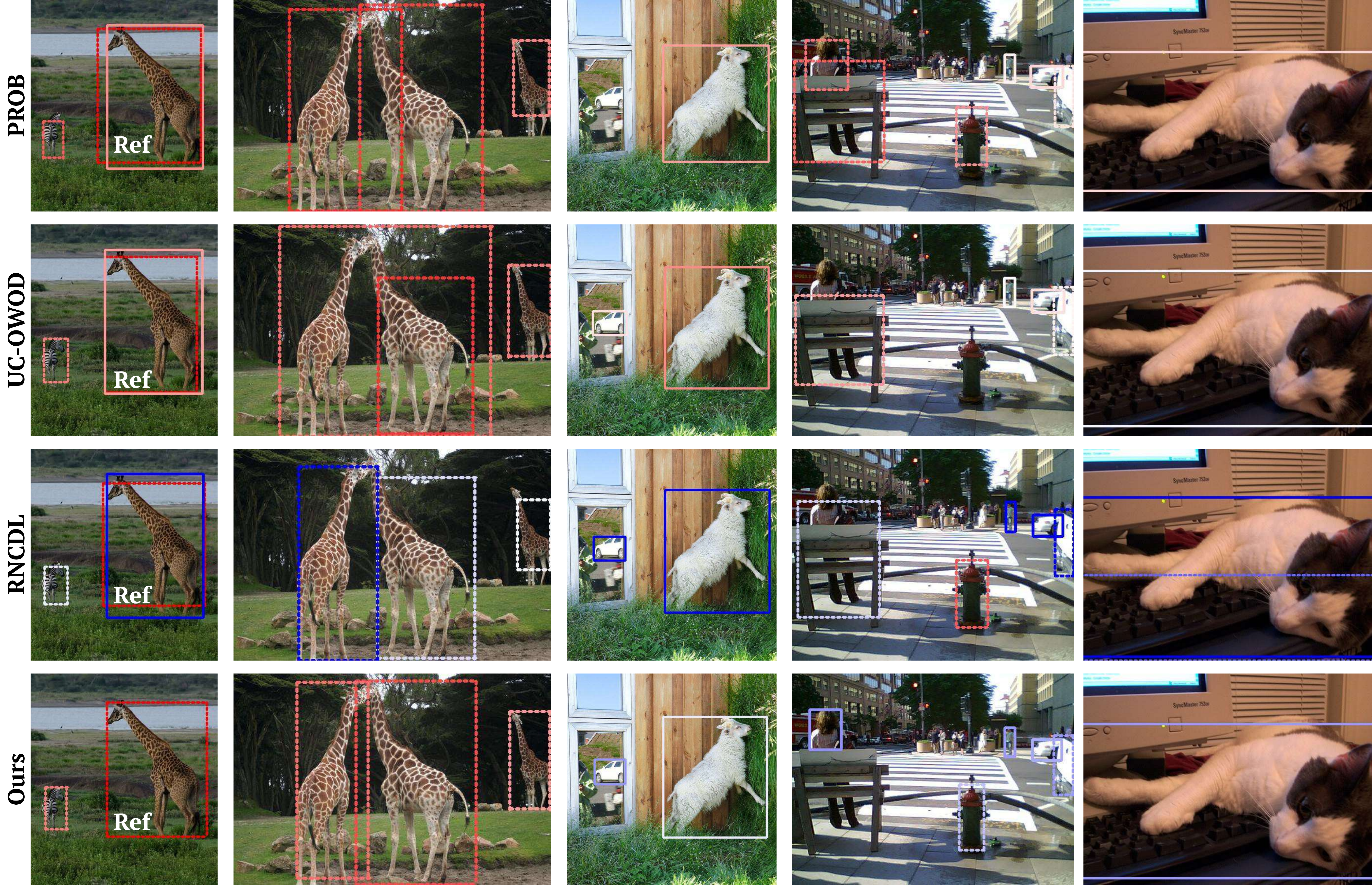}
\end{center}
\caption{
\textbf{Additional Qualitative Results on OWOD split.} We visualize the detection results of our model along with the instance embedding similarity with reference proposal. Red indicates high similarity, while blue represents high dissimilarity. Our method accurately detects both the known and unknown objects while simultaneously capturing fine-grained semantic relationships between proposals. For example, a \textit{giraffe} is considered more similar to a \textit{sheep} than to a \textit{fire hydrant}. In contrast, other approaches fail to capture such detailed semantic relationships.
}
\label{fig:sup_sim_owod}
\vspace{-0.1in}
\end{figure*}

\hspace{0.15in}Fig.~\ref{fig:sup_sim_owod} illustrates the results of inter-proposal relationships on the OWOD split. The embedding of \textit{giraffe} is used as reference. Our method considers other giraffe as highly similar, sheep as moderately similar, and fire hydrant as dissimilar, which reflects their semantic relationship. This demonstrates that our method can simultaneously detect unknown object and extract semantically rich feature from objects. \textit{PROB}, on the other hand,  considers all the proposals as highly similar since they are not trained to distinguish different unknown objects, but to learn a generalizable concept of object. Both \textit{UC-OWOD} and \textit{RNCDL} which learns instance embedding with self-supervision struggles to learn semantic relationship between objects. For more comprehensive comparison of the feature space, refer to Fig.~\ref{fig:tsne}.

\begin{figure*}[t]
\begin{center}
\includegraphics[width=1.0\linewidth]{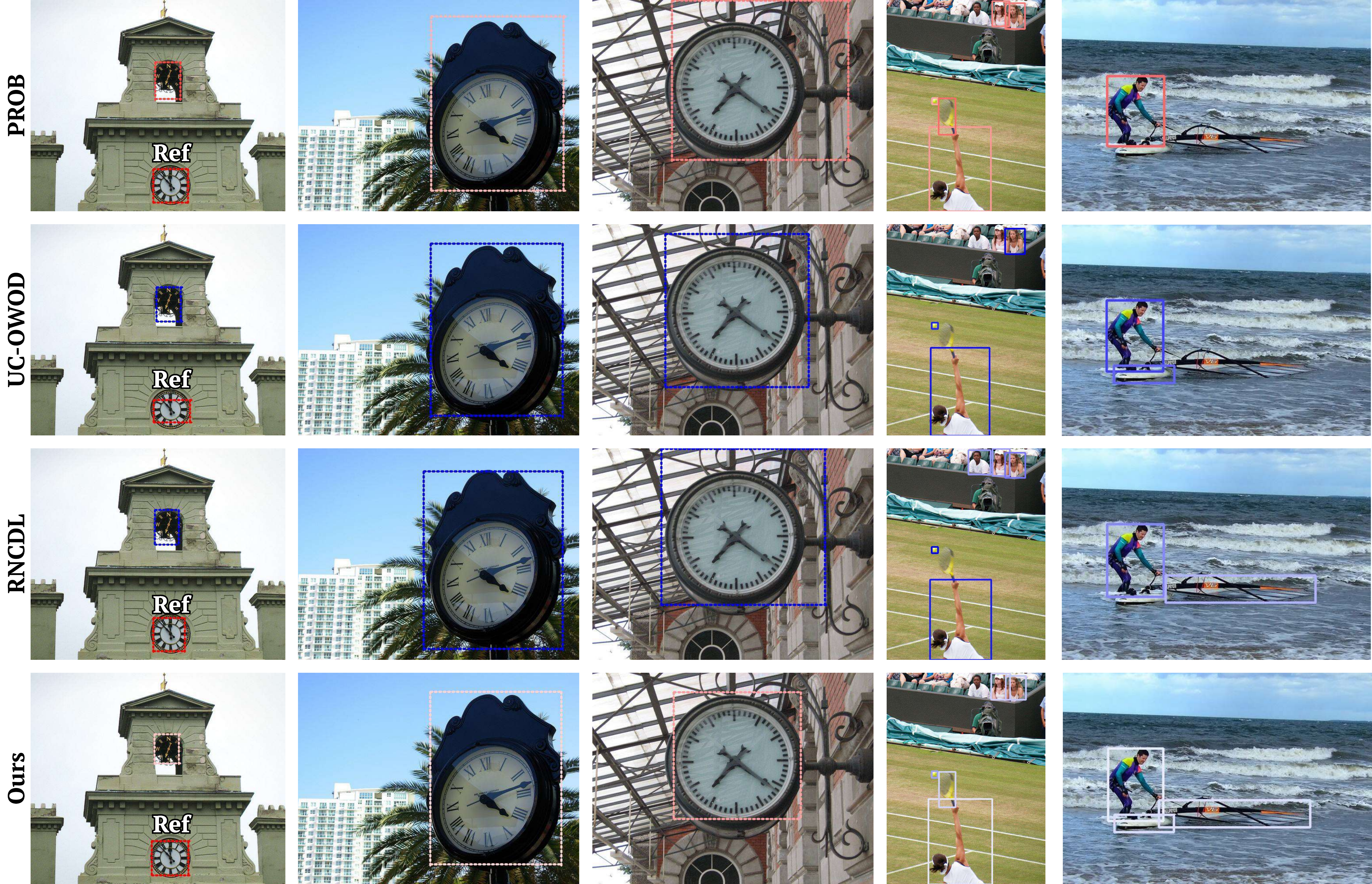}
\end{center}
\caption{
\textbf{Additional qualitative results on Unknown-Unknown split.} We visualize the detection results of our model along with the instance embedding similarity with the reference proposal of the \textit{clock}. Red indicates high similarity, while blue represents high dissimilarity. Even for the \textit{unknown-unknown}s that were not observed during training, our method accurately detects both the known and unknown objects while simultaneously capturing fine-grained semantic relationships between proposals. For example, a \textit{clock} is considered more similar to other \textit{clock} than to a \textit{tennis racket}. In contrast, other approaches fail to capture such detailed semantic relationships.
}
\label{fig:sup_sim_unk}
\vspace{-0.1in}
\end{figure*}

Fig.~\ref{fig:sup_sim_unk} presents the results of inter-proposal relationships on the \textit{Unknown-Unknown split}. Using the embedding of unknown \textit{clock} as reference, our method successfully captures the semantic relationship between objects even though they never observed \textit{clock} during training. For example, that other \textit{clock}s are highly similar, while \textit{person} or \textit{tennis racker} are dissimilar. Other methods suffers from learning such relationships correctly. This demonstrates that by distilling the rich feature through relaxed contrastive learning, our method learns a generalizable feature space of instance embeddings.

\subsection{Qualitative Results of Open-World Tracking.}
\begin{figure*}[t]
\begin{center}
\includegraphics[width=1.0\linewidth]{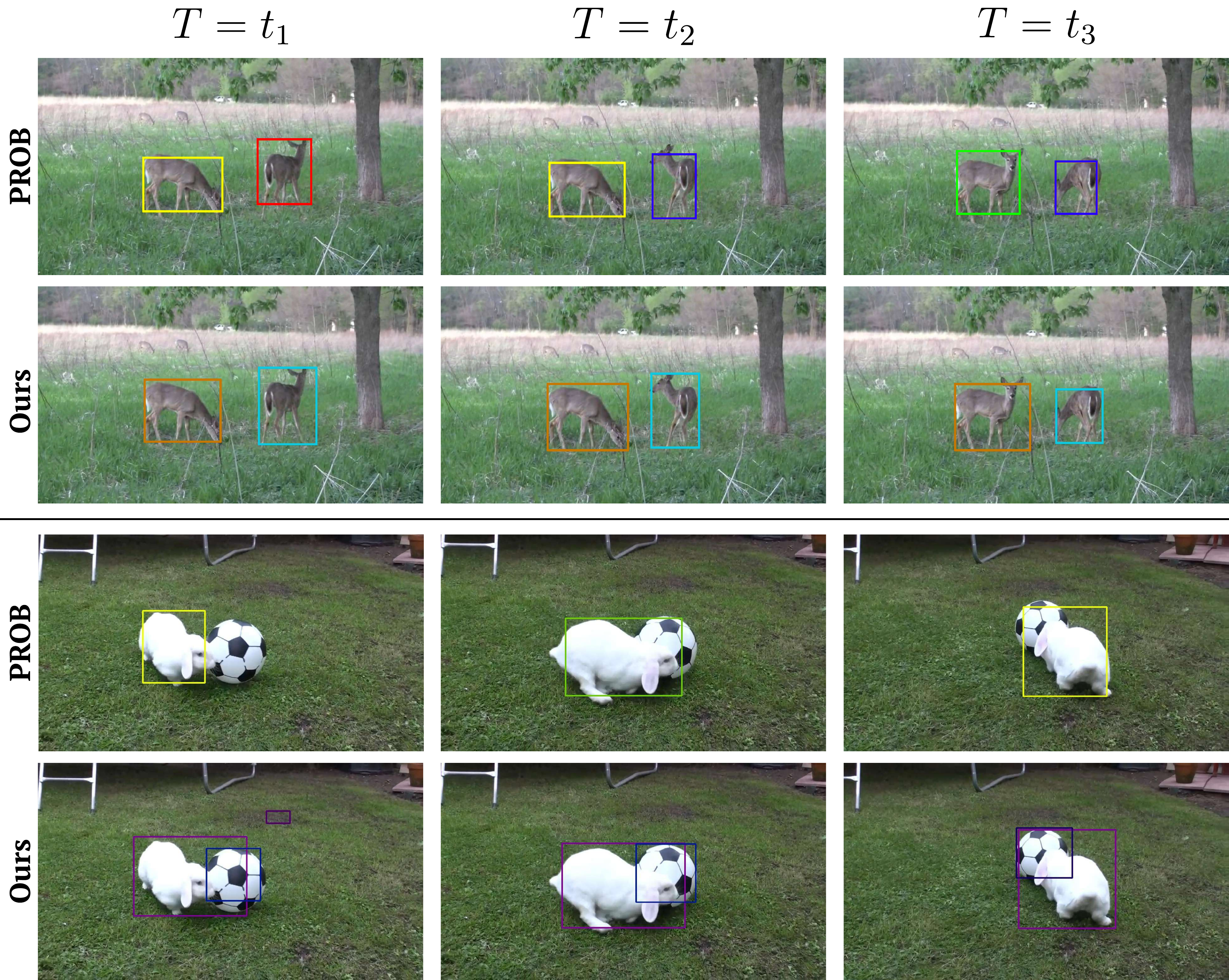}
\end{center}
\caption{
\textbf{Additional Qualitative Results of Open-World Object Tracking.} Bounding box color represents track IDs, with all the \textit{deer}, \textit{rabbit} and \textit{balls} belonging to unknown classes. Our method successfully tracks the unknown objects by learning semantically rich instance embeddings, used to compute feature similarity between inter-frame proposals.
}
\label{fig:sup_tracking}
\vspace{-0.1in}
\end{figure*}

\hspace{0.15in} We present additional results from our open-world tracking experiment, conducted using the detector trained on Task 1 of the OWOD split. The results are shown in Fig.~\ref{fig:sup_tracking}. Despite continuous shape changes in the \textit{deer} and \textit{rabbit}, the tracker built on our proposals and features successfully tracks the object. In contrast, when using \textit{PROB} outputs, the tracker struggles to assign proposals correctly as object shapes change. Additionally, \textit{PROB} fails to detect the unknown \textit{ball}. These results demonstrate that our method effectively extracts semantically rich features and detects unknown objects, improving open-world object tracking.


\begin{figure*}[t]
\begin{center}
\includegraphics[width=1.0\linewidth]{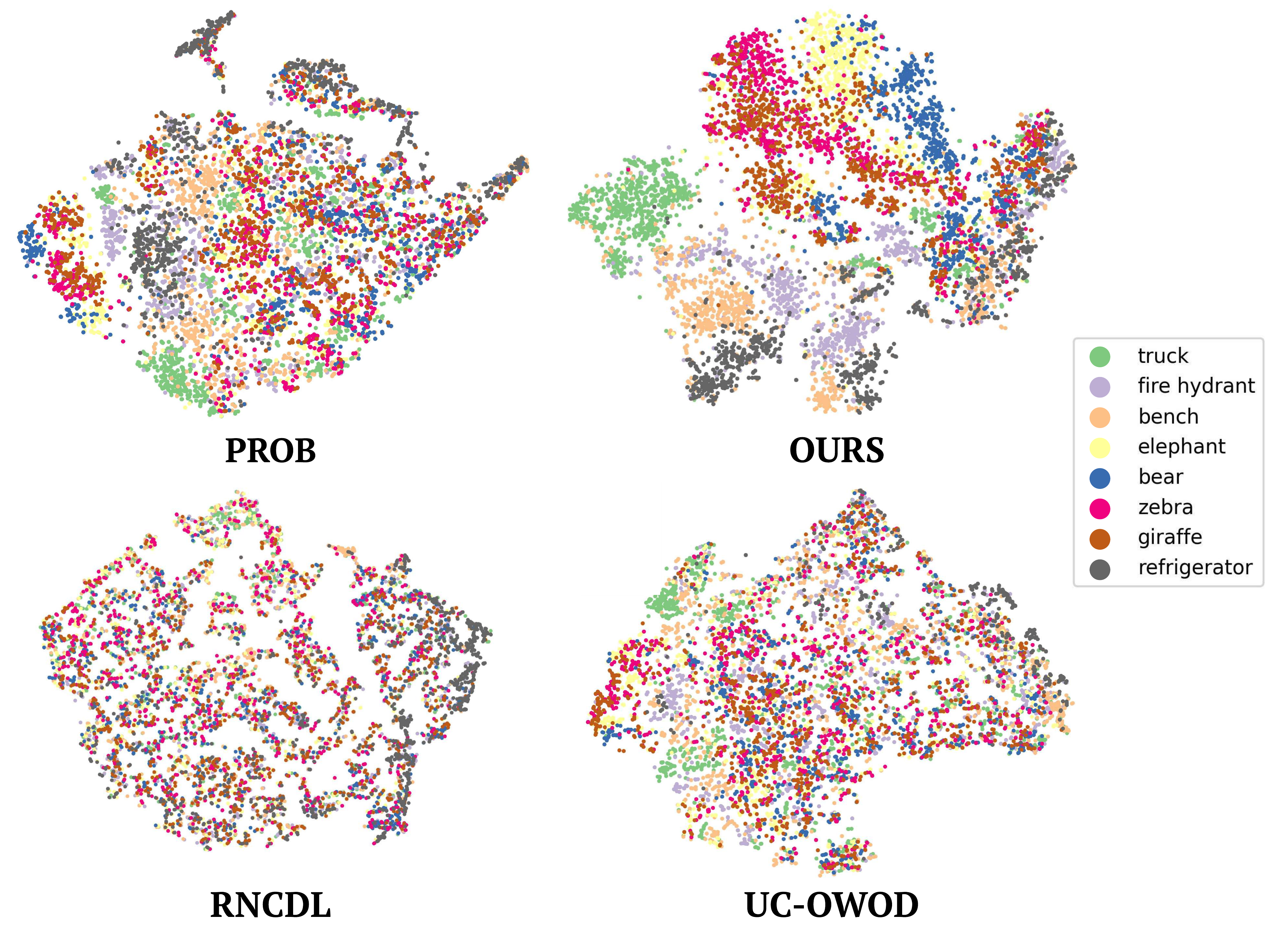}
\end{center}
\caption{
\textbf{Additional Qualitative Results on Inter-proposal Relationships.} We visualize the inter-proposal similarity of eight unknown classes. Our model groups proposals of the same class more compactly and allows similar animal classes to share a close feature space. In contrast, \textit{RNCDL} and \textit{UC-OWOD} exhibit a less effective feature space than \textit{PROB}.
}
\label{fig:tsne}
\vspace{-0.1in}
\end{figure*}

\begin{algorithm*}
\caption{Embedding Transfer Module (ETM) within a Training Iteration}
\label{alg:etm}
\begin{algorithmic}
\State \textbf{Input:} Image batch $\{I\}$, Ground truth $\{y\}$, Model $M$, DINOv2 model $M_{DINO}$
\State \textbf{Hyperparameters:} Margin $\delta$, Kernel bandwidth $\sigma$

\vspace{0.5em}
\State $\{B_{pred}\}, \{Q_{emb}\} \gets M(\{I\})$ \Comment{Get all predictions and query embeddings}

\vspace{0.5em}
\State $\{b_k\} \gets \text{HungarianMatcher}(\{B_{pred}\}, \{y\})$  \Comment{Get known proposals}
\State $\{b_u\} \gets \text{Top-k ObjectnesswithRefinement}(\{B_{pred}\})$  \Comment{Get top-k unknown proposals from URM}
\State $B \gets \{b_k\} \cup \{b_u\}$ 
\State $Q_{B} \gets \text{GetCorrespEmbeddings}(Q_{emb}, B)$ 

\vspace{0.5em}
\Statex \textit{Step 1: Get Source and Instance Embeddings}
\State $Z \gets \text{MLP}(Q_{B})$ 
\State $F \gets M_{DINO}(\{I\})$ \Comment{Extract DINOv2 feature map for the image batch. (A one-time pre-processing step.)}
\State $S \gets \text{AvgPool}(F, B)$  \Comment{Average Pool DINO embeddings from each proposals to obtain source embeddings}

\vspace{0.5em}
\Statex \textit{Step 2: Compute Pairwise Source Similarity (In practice, implemented using vectorized operations for efficiency.)}
\State $N = |S|$
\For{$i = 1$ to $N$, $j = 1$ to $N$}
        \State $w_{ij} \gets \exp\left(-\frac{\|s_i - s_j\|_2^2}{\sigma}\right)$ \Comment{ $\sigma$ : Gaussian kernel, $w_{ij}$ : Pairwise Source Similarity}
\EndFor

\vspace{0.5em}
\Statex \textit{Step 3: Compute Pairwise Instance Distance}
\For{$i = 1$ to $N$, $j = 1$ to $N$}
        \State $d_{ij} \gets \|z_i - z_j\|_2$  \Comment{$d_{ij}$: Pairwise Euclidean distance}
\EndFor

\vspace{0.5em}
\Statex \textit{Step 4: Compute Relaxed Contrastive Loss}

\State $\mathcal{L}_{et} =
\frac{1}{N}\sum_{i=1}^N
\sum_{j=1}^N
\left[
\mathbf{w}_{ij}\mathbf{d}_{ij}^2 +(1-\mathbf{w}_{ij})
\left[\delta - \mathbf{d}_{ij}
\right]_{+}^2
\right]$ \Comment{ $\delta$ : margin}

\vspace{0.5em}
\State \textbf{Return} $\mathcal{L}_{et}$
\end{algorithmic}
\end{algorithm*}

\end{document}